%% file: draft.tex
\documentclass[acmsmall,nonacm,screen]{acmart}

\setcopyright{none}
\acmJournal{PACMPL}
\acmYear{2020}
\copyrightyear{2020}
\acmVolume{4}
\acmNumber{OOPSLA}
\acmArticle{}
\acmMonth{11}
\acmPrice{}
\acmDOI{10.1145/3428230}

\input{packages}

\input{genericmacros}

\input{specificmacros}

\input{acronyms}

\begin{document}

\title[Neural Reverse Engineering of Stripped Binaries using Augmented Control Flow Graphs]{Neural Reverse Engineering of Stripped Binaries \\ using Augmented Control Flow Graphs}

\author{Yaniv David}
\affiliation{
  \institution{Technion}            %
  \country{Israel}                    %
}
\email{yanivd@cs.technion.ac.il}          %

\author{Uri Alon}
\affiliation{
  \institution{Technion}            %
  \country{Israel}                    %
}
\email{urialon@cs.technion.ac.il}          %
\author{Eran Yahav}
\affiliation{
  \institution{Technion}            %
  \country{Israel}                    %
}
\email{yahave@cs.technion.ac.il}          %

\input{abstract.tex}

\maketitle

\input{intro}

\input{overview}

\input{background}

\input{representation}

\input{model}

\input{eval}

\input{related}

\input{conclusion}
\input{ack}

\appendix
\input{appendix.tex}

\clearpage

\bibliography{references}

\end{document}

%% file: packages.tex
\usepackage{microtype}

\usepackage[nolist,nohyperlinks]{acronym}
\usepackage{booktabs}
\usepackage{subcaption}
\usepackage{float}
\usepackage{listings}
\usepackage[linesnumbered,ruled,vlined]{algorithm2e}
\usepackage{hyperref}
\usepackage{paralist}
\usepackage{newtxmath}

\usepackage{graphicx}
\usepackage{array}

\usepackage{lipsum}
\usepackage{multicol}
\usepackage{multirow}
\usepackage{wrapfig}

\usepackage{enumitem} %

\newenvironment{framed}{}

\usepackage[capitalize]{cleveref}

%% file: genericmacros.tex
\usepackage{ifthen}

\newcommand{\AWK}[1]{ {\color{red} \bf #1} }

\newcommand{\TODO}[1]{}

\newcommand{\BF}[1]{\textbf{#1}}

\newcommand{\todolow}[1]{}

\newcommand{\COMMENT}[1]{}
\newcommand{\ignore}[1]{}
\newcommand{\ignoreRC}[1]{}
\newcommand{\ie}{i.e., }
\newcommand{\eg}{e.g., }

\ignore{ no need for these anymore, cref for the win:

}

\newboolean{TR}
\setboolean{TR}{false}
\ifthenelse{\boolean{TR}}{

\newcommand{\TrOnly}[1]{#1}
\newcommand{\SubOnly}[1]{}
\newcommand{\TrOnlyInFootnote}[1]{#1}
\newcommand{\TrOnlyInTable}[1]{#1}}
{

\newcommand{\TrOnly}[1]{}
\newcommand{\SubOnly}[1]{#1}
\newcommand{\TrOnlyInFootnote}[1]{}
\newcommand{\TrOnlyInTable}[1]{}}

\newcommand{\para}[1]{\vspace{3pt}\noindent\textbf{\textit{#1}}}
\newcommand{\scode}[1]{{\small \texttt{#1}}}

\definecolor{airforceblue}{rgb}{0.36, 0.54, 0.66}
\definecolor{awesome}{rgb}{1.0, 0.13, 0.32}
\definecolor{blue(pigment)}{rgb}{0.2, 0.2, 0.6}
\definecolor{britishracinggreen}{rgb}{0.0, 0.26, 0.15}
\definecolor{cadmiumgreen}{rgb}{0.0, 0.42, 0.24}
\definecolor{carminered}{rgb}{1.0, 0.0, 0.22}
\definecolor{electricindigo}{rgb}{0.44, 0.0, 1.0}
\definecolor{mygray}{rgb}{0.5,0.5,0.5}
\hypersetup{colorlinks=true,linkcolor=black, citecolor=black, urlcolor=black}

\newcommand{\cbox}[1]{\textcircled{\raisebox{0pt}{\footnotesize{#1}}}}%

%% file: specificmacros.tex
\newcommand{\YD}[2]{ {\color{red} #1 (\textbf{YD}: \emph{#2})} }

\newcommand{\oursite}{\url{https://github.com/tech-srl/Nero}}

\definecolor{uricolor}{HTML}{0F9D58}
\newcommand{\uri}[1]{{\color{uricolor} (@uri: #1)}}
\definecolor{uricolorV}{HTML}{0AAAAA}

\definecolor{yanivcolor}{HTML}{0A9D58}

\definecolor{yanivcolorV}{HTML}{0BBBBB}

\definecolor{ourpurple}{HTML}{800080}
\definecolor{ourgreen}{HTML}{228B22}

\newcommand{\lsyn}{\lbrack\!\lbrack}
\newcommand{\rsyn}{\rbrack\!\rbrack}
\newcommand{\semp}[1]{\lsyn #1 \rsyn}

\newcommand{\tool}{\scode{Nero}}
\newcommand{\tools}{\scode{Nero\tiny{ }}}

\newcommand{\toolL}{\tool\scode{-LSTM}}

\newcommand{\toolT}{\tool\scode{-Transformer}}

\newcommand{\toolsTL}{\tools \scode{Transformer$\rightarrow$LSTM}}

\newcommand{\toolG}{\tool\scode{-GNN}}

\newcommand{\debin}{Debin}
\newcommand{\debins}{Debin~}
\newcommand{\dire}{DIRE}
\newcommand{\dires}{DIRE~}

\newcommand{\ScoreLSTM}{21.72}
\newcommand{\ScoreGNero}{45.53}
\newcommand{\ScoreDebin}{33.66}
\newcommand{\NeroAugmentImprove}{20\%}
\newcommand{\NeroVSText}{100\%}
\newcommand{\NeroVSDebin}{35\%}
\newcommand{\NeroVSDIRE}{28\%}

\lstdefinelanguage
   [x86_64]{Assembler}     %
   [x86masm]{Assembler} %
   {morekeywords={CDQE,CQO,CMPSQ,CMPXCHG16B,JRCXZ,LODSQ,MOVSXD, %
                  POPFQ,PUSHFQ,SCASQ,STOSQ,IRETQ,RDTSCP,SWAPGS, %
                  rax,rdx,rcx,rbx,rsi,rdi,rsp,rbp,r9,r9d,r12,   %
                  r12d,r12b,r13,r13d,r14,r14d,r15,r15d,         %
                  zf, of, pf,                                   %
                  load, store, icmp,  %
                  }} %

\lstdefinestyle{x86_64asm}{
        basicstyle=\ttfamily,
        language=[x86_64]Assembler,
        emph={},
        emphstyle={\underbar},
        numbers=none,
        numberfirstline=false,
        numberstyle=,
        escapeinside={(*@}{@*)}
}

\lstdefinestyle{AArch64asm}{
        basicstyle=\ttfamily\scriptsize,
        language=[AArch64]Assembler,
        emph={},
        emphstyle={\underbar},
        numbers=none,
        numberfirstline=false,
        numberstyle=\tiny,
        escapeinside={(*@}{@*)}
}

\lstdefinelanguage
   [MIPS]{Assembler}     
   [x86masm]{Assembler} 
   {morekeywords={jalr, move, li, lw, bne, addiu, jal, beq, lui, %
                  gp, a0, a1,a2,s2,s4,s5,s6,v0,v1,t9  %
                  }} %

\lstdefinestyle{MIPSasm}{
        basicstyle=\ttfamily\scriptsize,
        language=[MIPS]Assembler,
        emph={},
        emphstyle={\underbar},
        numbers=left,
        numberfirstline=false,
        numberstyle=\tiny,
        escapeinside={(*@}{@*)}
}

\lstdefinestyle{LLVMir}{
        basicstyle=\ttfamily\scriptsize,
        language=[LLVM]Assembler,
        emph={},
        emphstyle={\underbar},
        numbers=none,
        numberfirstline=false,
        numberstyle=\tiny,
        escapeinside={(*@}{@*)}
}

\lstdefinestyle{x86_64asm-small}{
        basicstyle=\ttfamily\tiny,
        language=[x86_64]Assembler,
        emph={},
        emphstyle={\underbar},
        numbers=left,
        numberstyle=\tiny,
        escapeinside={(*@}{@*)}
}

\lstdefinestyle{AArch64asm-small}{
        basicstyle=\ttfamily\tiny,
        language=[AArch64]Assembler,
        emph={},
        emphstyle={\underbar},
        numbers=left,
        numberstyle=\tiny,
        escapeinside={(*@}{@*)}
}

%% file: acronyms.tex
\begin{acronym}
\acro{ABI}{application binary interface}
\acro{API}{application program interface}
\acro{AST}{abstract syntax tree}
\acro{AUC}{area under (the) curve}

\acro{BB}{basic block}

\acro{CROC}{concentrated ROC}
\acro{CFG}{control-flow graph}
\acro{CU}{compilation unit}
\acro{CVE}{common vulnerabilities (and) exposures}
\acro{CPU}{central processing unit}

\acro{DAG}{directed acyclic graph}
\acro{DNN}{deep nural network}

\acro{ELF}{executable and linkable format}

\acro{GOT}{global offset table}
\acro{GNN}{graph neural network}
\acro{GCN}{graph convolutional network}

\acro{FP}{false positive}
\acro{FN}{false negative}
\acro{FOL}{first order logic}

\acro{HTTP}{the hypertext transfer protocol}

\acro{IL}{intermediate language}
\acro{IOT}{internet of things}
\acro{IP}{intellectual property}
\acro{IR}{intermediate representation}
\acro{ISA}{instruction set architecture}
\acro{IVL}{intermediate verification language}

\acro{LCS}{longest common subsequence}
\acro{LSTM}{long short-term memory network}

\acro{ML}{machine language}

\acro{NLP}{natural language processing}
\acro{NMT}{neural machine translation}

\acro{OS}{operating system}
\acro{OOV}{out-of-vocabulary}

\acro{PDG}{program dependence graph}
\acro{PIC}{position independent code}

\acro{TP}{true positive}
\acro{TN}{true negative}

\acro{RE}{reverse engineering}
\acro{ROC}{receiver operating characteristic}
\acro{RNN}{recurrent neural network}

\acro{SSL}{secure sockets layer}
\acro{SSA}{single static assignment}
\acro{seq2seq}{sequence-to-sequence}
\acro{SSA}{single static assignment}
\end{acronym}

\acrodefplural{CU}[CUs]{compilation units}
\acrodefplural{CFG}[CFGs]{control flow graphs}
\acrodefplural{CVE}[CVEs]{common vulnerabilities and exposures}

\acrodefplural{ISA}[ISAs]{architecture instruction sets}

\acrodefplural{LSTM}[LSTMs]{long short-term memory networks}
\acrodefplural{GNN}[GNNs]{graph neural networks}

\acrodefplural{OS}[OSs]{operating systems}

\acrodefplural{RNN}[RNNs]{recurrent neural networks}

%% file: abstract.tex
\begin{abstract}
We address the problem of reverse engineering of stripped executables, which contain no debug information. This is a challenging problem because of the low amount of syntactic information available in stripped executables, and the diverse assembly code patterns arising from compiler optimizations.

We present a novel approach for predicting procedure names in stripped
executables. Our approach combines static analysis with neural models. The
main idea is to use static analysis to obtain augmented representations of
\emph{call sites}; encode the structure of these call sites using the \ac{CFG}
\acresetall and finally, generate a target name while attending to these call
sites. We use our representation to drive graph-based, LSTM-based and
Transformer-based architectures. 

Our evaluation shows that our models produce predictions that are difficult
and time consuming for humans, while improving on existing methods by
$\NeroVSDIRE$ and by $\NeroVSText$ over state-of-the-art neural textual models
that do not use any static analysis. Code and data for this evaluation are
available at \oursite{}.

\end{abstract}

%% file: intro.tex
\section{Introduction}\label{Sec@Intro}

\ignore{
In early 2011, Symantec published its findings and analysis about ``Stuxnet''~\cite{StuxnetDossier}, one of the most complex and advanced malware the world has ever seen. In their timeline they state that: ``Thousands of man hours [were invested] analyzing 500 kilobytes of code (...), 40 sleepless nights, and 10 weekends later, $95\%$ of code [has been] reverse engineered''~\cite{StuxnetTimeline}. Almost a decade has passed, and the battle between malware authors and security researchers is far from over.
}

Reverse engineering (RE) \acused{RE} of executables has a variety of
applications such as improving and debugging legacy programs. Furthermore, it is
crucial to analyzing malware. Unfortunately, it is a hard skill to learn, and it
takes years to master. Even experienced professionals often have to invest
long hours to obtain meaningful results. The main challenge is to understand
how the different ``working parts'' inside the executable are meant to
interact to carry out the objective of the executable.  A human
reverse-engineer has to \emph{guess}, based on experience, the more
interesting procedures to begin with, follow the flow in these procedures, use
inter-procedural patterns and finally, piece all these together to develop a
global understanding of the purpose and usage of the inspected executable. 

Despite great progress on disassemblers~[\citeauthor{IDAPRO, RADAR}], static
analysis frameworks~\cite{KatzClass,TIE} and similarity
detectors~\cite{GITZ,Pewny}, for the most part, the reverse engineering process remains 
manual. 

Reviewing source code containing meaningful names for procedures can reduce
human effort dramatically, since it saves the time and effort of looking at
some procedure bodies~\cite{alon2019code2vec,host2009debugging,
fowler1999refactoring,JacobsonRM11}. Binary executables are usually stripped,
\ie the debug information containing procedure names is removed entirely.

As a result of executable stripping, a major part of a reverse engineer's work is
to manually label procedures after studying them.
\citet{votipka2020observational} detail this process in a user study of
reverse engineers and depict their reliance on internal and external
procedure names throughout their study. 

\ignore{
 ``when P08V [a participant] saw a function named \scode{httpd\_ipc\_init}, he recognized this might introduce a vulnerability(...)''. 
}

In recent years, great strides have been made in the analysis of source code using
learned models from automatic inference of variables and types
\cite{raychev2015jsnice,bielik16phog, alon2018general,
bavishi2018context2name, allamanis2017learning} to bug
detection~\cite{pradel2018deepbugs, rice2017detecting}, code
summarization~\cite{allamanis16convolutional, alon2019code2vec,
alon2018code2seq}, code retrieval \cite{sachdev2018retrieval,
allamanis2015bimodal} and even code generation~\cite{murali2018bayou,
brockschmidt2018generative, lu2017data, alon2019structural}. However, all of
these address high-level and syntactically-rich programming languages such as
Java, C\# and Python. None of them address the unique challenges present in
executables.

\para{Problem definition}
Given a nameless assembly procedure $\mathcal{X}$ residing in a stripped
(containing no debug information) executable, our goal is to predict a likely
and descriptive name $\mathcal{Y}=y_1...,y_m$, where $y_1...,y_m$ are the
subtokens composing $\mathcal{Y}$. Thus, our goal is to model
$P\left(\mathcal{Y} \mid \mathcal{X}\right)$. For example, for the name
$\mathcal{Y}=$ \scode{create\_server\_socket}, the subtokens $y_1...,y_m$ that
we aim to predict are \scode{create}, \scode{server} and \scode{socket},
respectively.

The problem of predicting a meaningful name for a given procedure can be
viewed as a translation task -- translating from assembly code to natural
language. While this high-level view of the problem is shared with previous
work (e.g.,~\cite{allamanis16convolutional, alon2019code2vec,
alon2018code2seq}), the technical challenges are vastly different due to the
different characteristic of \emph{binaries}.

\input{code/shuffled_calls_figure}

\para{Challenge 1: Little syntactic information and token coreference}
Disassembling the code of a stripped procedure results in a sequence of
instructions as the one in \cref{Fig@ASM@Dissassmby}. These instructions are
composed from a mnemonic (\eg \scode{mov}), followed by a mix of register names and
alphanumeric constants. These instructions lack any information regarding the
variable types or names that the programmer had defined in the high-level
source code. Registers such as \scode{rsi} and \scode{rax} (lines \thersiIC, \theraxC~in
\cref{Fig@ASM@Dissassmby}) are referenced in different manners, \eg
\scode{rax} or \scode{eax} (lines \thersiIC, \thersiIIC), and used
interchangeably to store the values held by different variables. This means
that the presence of a register name in an instruction carries little
information. The same is true for constants: the number $4$ can be an offset to a
stack variable (line \theraxC), an Enum value used as an argument in a
procedure call (line \thefourC) or a jump table index.

Learning descriptive names for procedures from such a low-level stripped
representation is a challenging task. A na\"ive approach, where the flat
sequence of assembly instructions is fed into a \ac{seq2seq}
architecture~\cite{luong15,vaswani2017attention}, yields imprecise results
($\ScoreLSTM$~F1 score), as we show in \cref{Sec@Eval}.

\para{Challenge 2: Long procedure names}  Procedure names in compiled C code
are often long, as they encode information that would be part of a typed
function signature in a higher-level language (e.g.,  {\small
\scode{AccessCheckByTypeResultListAndAuditAlarmByHandleA}} in the Win32 API).
Methods that attempt to directly predict a full label as a single word from a
vocabulary will be inherently imprecise.

\para{Our approach} We present a novel representation for binary procedures
specially crafted to allow a neural model to generate a descriptive name for a
given stripped binary procedure. To construct this representation from a
binary procedure we:  \begin{enumerate}  \item Build a control-flow graph (\ac{CFG}) from the
disassembled binary procedure input. \item Reconstruct a call-site-like
structure for each
call instruction present in the disassembled code. \item Use pointer-aware slicing to augment these call
sites by finding concrete values or approximating abstracted values. 
\item Transform the \ac{CFG} into an augmented call sites
graph.\end{enumerate} This representation is geared towards addressing the
challenge of analyzing non-informative and coreference riddled disassembled
binary code (challenge 1).

In \cref{Sec@Model}, we explain how this representation is combined with
graph-based \cite{kipf2017semi}, \ac{LSTM}-based \cite{sutskever2014sequence,
luong15} and Transformer-based \cite{vaswani2017attention} neural
architectures. These architectures can decode long \ac{OOV} predictions to address the challenge of predicting long names for binary procedures (challenge 2).

This combination provides an \emph{interesting and powerful balance} between the
program-analysis effort required to obtain the representation from binary
procedures, and the effectiveness of the learning model.

\ignore{

In \cref{Sec@Eval} we show that our approach outperforms
existing approaches and na\"ive textual application of \acp{LSTM} and
Transformers on the raw disassembled binary code. 

In this work, we propose a novel representation for ~\emph{stripped,
optimized, compiled assembly code}. This representation can be used in any
neural architecture. To the best of our knowledge, this is the first work to
leverage deep learning for reverse engineering procedure names in the
realistic settings of \YD{stripped}{Maybe go back here and say dire was not
fully stripped} \uri{The fact that they didn't evaluate on stripped binaries is not important in this section. What is important (here or in the previous paragraph) --  is what limits them from addressing fully-stripped and fully-optimized binaries?  -> i have no idea why they didn't optimize} and optimized binaries.
}

\para{Existing techniques} \debin{} \cite{he2018debin} is a non-neural model for reasoning about
binaries. This model suffers from inherent sparsity and can only predict full
labels encountered during training.

\dire{} \cite{lacomis2019dire} is a neural model for predicting variable
names. \dire{} uses the textual sequence (C code) created by an external
black-box decompiler (\citeauthor{HexRays}), accompanied by the \ac{AST} to create a
hybrid \ac{LSTM} and \ac{GNN} based model. \TODO{ i need something to say here \uri{"However, the reliance of \dire{} on a black-box decompiler allows it to work well only in \emph{non-optimized binaries}; optimized binaries are usually not decompiled in a readable or meaningful way."}}

In \cref{Sec@Eval}, we show that our approach outperforms \dires and \debins with
relative F1 score improvements of $\NeroVSDIRE$ and $\NeroVSDebin$,
respectively.

\para{Main contributions} The main contributions of this work are:
\begin{itemize} 

\item A novel representation for binary procedures. This representation is
based on augmented call sites in the \ac{CFG} and is tailor-made for procedure
name prediction.

\item This representation can be used in a variety of neural architectures
such as \acp{GNN}, \acp{LSTM} and Transformers (\cref{Sec@Repr}). 

\item \tool{}\footnote{The code for \tools and other resources are publicly available at \oursite} , A framework for binary name prediction. \tool{} is composed of:
\begin{inparaenum}[(i)] \item a static binary analyzer for producing the
augmented call sites representation from stripped binary procedures and \item
three different neural networks for generating and evaluating name prediction\end{inparaenum}.

\item An extensive evaluation of \tool{}, comparing it with \debin, \dires and
state-of-the-art \ac{NMT} models (\cite{luong15, vaswani2017attention}). \tool{} achieves an F1 score of $\ScoreGNero$ in predicting procedure
names within GNU packages, outperforming all other existing techniques.

\item A thorough ablation study, showing the importance of the different components in our approach (\cref{SSec@Eval@Ablation}).

\end{itemize}

%% file: code/shuffled_calls_figure.tex
\newcounter{CShuffled}

\begin{figure}
\centering
\begin{tabular}{c|c|c}

\begin{subfigure}{0.24\textwidth}
\begin{framed}
\centering
{
\begin{framed}
    \small
    \raggedright
...                        \\ \refstepcounter{CShuffled}
\newcounter{rsiIC} \setcounter{rsiIC}{\theCShuffled}
\theCShuffled:  mov  rsi, rdi       \\ \stepcounter{CShuffled}
\theCShuffled:  mov  rdx, 16        \\ \stepcounter{CShuffled} 
\theCShuffled:  mov  rax, [rbp-58h] \\ \stepcounter{CShuffled}
\theCShuffled:  mov  rdi, rax       \\ \stepcounter{CShuffled}
\newcounter{ConnectlineC} \setcounter{ConnectlineC}{\theCShuffled}
\theCShuffled:  \textbf{call  connect} \\ \stepcounter{CShuffled}
\theCShuffled:  mov     edx, eax \\   \stepcounter{CShuffled}
\newcounter{raxC} \setcounter{raxC}{\theCShuffled}
\theCShuffled:  mov     rax, [rbp-4h] \\ \stepcounter{CShuffled}
...  \\
\theCShuffled:  mov  rax, [rbp-58h] \\ \stepcounter{CShuffled}
\theCShuffled:  mov  rdi, rax \\ \stepcounter{CShuffled}
\newcounter{fourC} \setcounter{fourC}{\theCShuffled}
\theCShuffled:  mov  r8,  4 \\ \stepcounter{CShuffled}
\theCShuffled:  mov  rdx, 10 \\ \stepcounter{CShuffled}
\newcounter{rsiIIC} \setcounter{rsiIIC}{\theCShuffled}
\theCShuffled:  mov  esi, 0 \\ \stepcounter{CShuffled}
\theCShuffled:  lea  rcx, [rbp-88h] \\ \stepcounter{CShuffled}
\newcounter{ssoC} \setcounter{ssoC}{\theCShuffled}
\theCShuffled: \textbf{call  setsockopt} \\ \stepcounter{CShuffled}
...
\end{framed}
}
\end{framed}
\end{subfigure}

&

\begin{subfigure}{0.24\textwidth}
\begin{framed}
\centering
{
\begin{framed}
    \small
    \raggedright
    1: \textbf{call socket} \\
    \quad call printf \\
    \quad call close \\
    3: \textbf{call connect} \\
    \quad call printf \\
    2: \textbf{call setsocketopt} \\
\end{framed}
}
\end{framed}
\end{subfigure}

&

\begin{subfigure}{0.51\textwidth}
\begin{framed}
\centering
\includegraphics[keepaspectratio, width=0.9\textwidth]{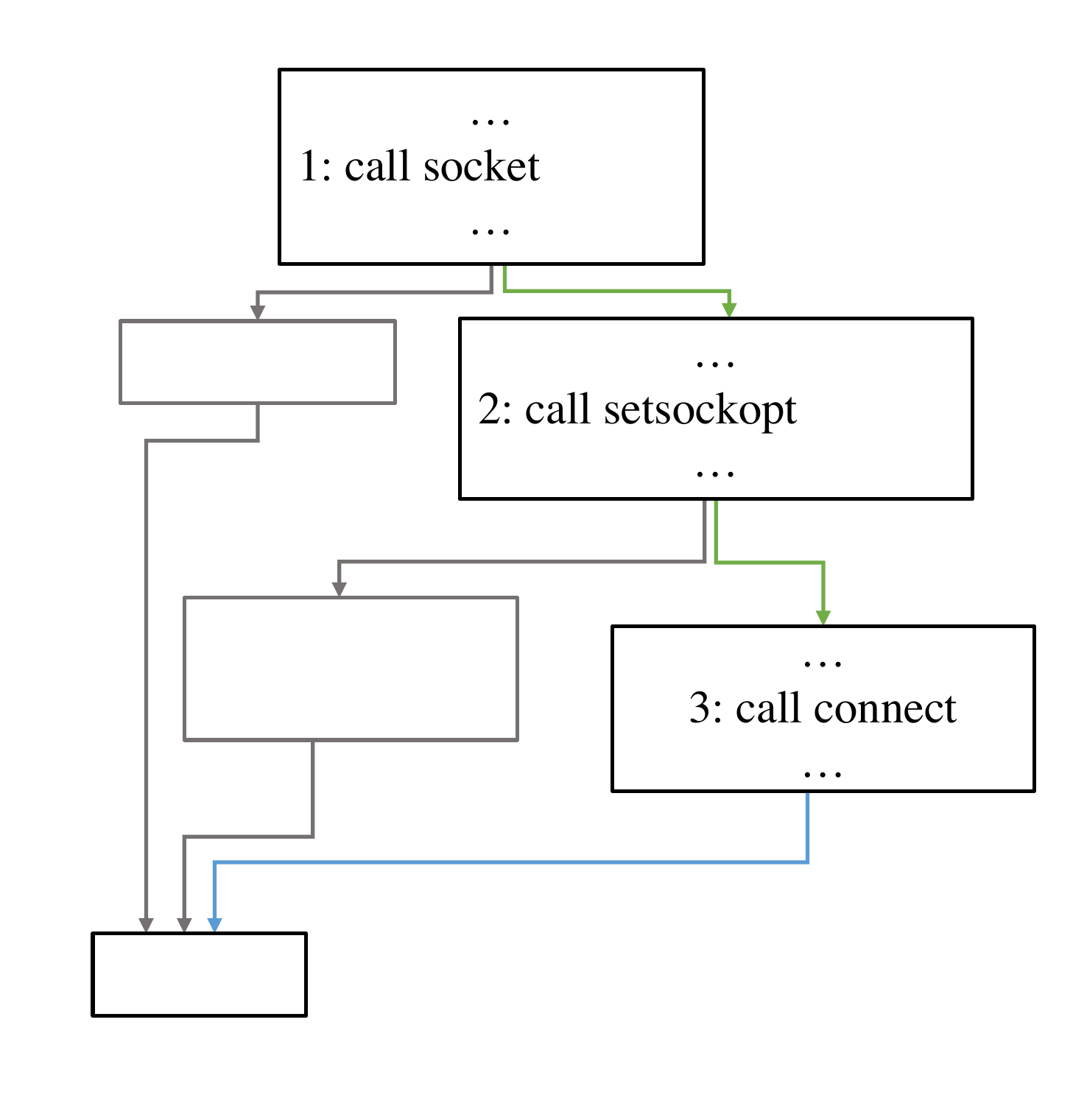}
\end{framed}
\end{subfigure}

\\

\begin{subfigure}{0.24\textwidth}
\begin{framed}
\caption{}
\label{Fig@ASM@Dissassmby}
\end{framed}
\end{subfigure}

&

\begin{subfigure}{0.24\textwidth}
\begin{framed}
\caption{}
\label{Fig@ASM@CallSequence}
\end{framed}
\end{subfigure}

&

\begin{subfigure}{0.5\textwidth}
\begin{framed}
\caption{}
\label{Fig@ASM@CFG}
\end{framed}
\end{subfigure}

\end{tabular}

\caption{(a) The result of disassembing an optimized binary procedure taken
from a stripped executable ; (b) The call instructions randomly placed in the
procedure code by the compiler ; (c) the call instructions are placed in their
correct \emph{call order} as they appear in the \ac{CFG}.} 
\label{Fig@ASM}
\end{figure}

%% file: overview.tex
\section{Overview} \label{Sec@Overview}

In this section, we illustrate our approach informally and provide an example to further explain the challenges in binary procedure name prediction. This example is based on  a procedure
from our test set that we simplified for readability purposes. While this
example is an Intel 64-bit Linux executable, the same process can be
applied to other architectures and operating systems.

\para{Using calls to imported procedures} Given an unknown binary procedure
$P$, which is stripped of debug symbols, our goal is to predict a meaningful
and likely name that describes its purpose. After disassembling $P$'s code,
the procedure is transformed into a sequence of assembly instructions. A
snippet from this sequence is shown in \cref{Fig@ASM@Dissassmby}. 

We draw our initial intuition from the way a human reverse engineer skims this
sequence of assembly instructions. 
The most informative pieces of information to understand what the code does are \emph{calls to procedures} whose names
are known because they \emph{can be statically resolved} and can not be easily stripped\footnote{See \cref{SSec@Eval@TestsetCreation} for more details about API name obfuscation, and how we simulated it in our evaluation.}. In our example, these
are \scode{call connect} (line $\theConnectlineC$) and \scode{call setsockopt}
(line $\thessoC$) in \cref{Fig@ASM@Dissassmby}. Resolving these names is
possible because these called procedures reside in libraries that are dynamically
linked to the executable, causing them to be imported (into the executable
memory) as a part of the \ac{OS} loading process. We further discuss imported procedure name resolution in \cref{Sec@Repr}. Calls to such imported
procedures are also called \ac{API} calls, as they expose an interface to these
libraries. We will refer to their names, \eg \scode{connect} and
\scode{setsockopt}, as API names.

In order to pass arguments when making these API calls, the calling convention
used by the \ac{OS} defines that these argument values are placed into
specific registers before the call instruction. The process of assigning into
these registers is shown in lines 1-4 and 8-13 of \cref{Fig@ASM@Dissassmby}.
Its important to note, while anti-\ac{RE} tools may try and obfuscate API
names, the \emph{values} for the arguments passed when calling these external
procedures must remain intact. 

In \cref{Sec@Eval} we discuss API name obfuscation and examine their effect on
the performance of different approaches to procedure name prediction.

\para{Putting calls in the right order} After examining the API calls of the
procedure, a human reverser will try to understand the \emph{order} in which
they are called at runtime. \Cref{Fig@ASM@CallSequence} shows all the \scode{call}
instructions in $P$'s disassembled code, in the semi-random order 
in which they were generated by the compiler. This order does not reflect any logical or
chronological order. For example, \begin{inparaenum}[(i)]  \item
the call to \scode{socket}, which is part of the setup, is interleaved with
\scode{printf} used for error handling; and  \item the calls in this
sequence are randomly shuffled in the assembly, i.e., \scode{connect}
appears before \scode{setsocketopt} \end{inparaenum}.

To order the API calls correctly, we construct a \ac{CFG} for $P$. A \ac{CFG}
is a directed graph representing the control flow of the procedure, as
determined by jump instructions. \Cref{Fig@ASM@CFG} shows a \ac{CFG} created
from the dissembled code in \cref{Fig@ASM@Dissassmby}. For readability only,
the API calls are presented in the \ac{CFG} nodes (instead of the full
instruction sequence). By observing the \ac{CFG}, we learn all possible runtime
paths, thus approximating all potential call sequences.

A human reverse-engineer can follow the jumps in the \ac{CFG} and figure out the
order in which these calls will be made:
$\pi=$~\scode{socket}$\rightarrow$\scode{setsockopt}$\rightarrow$\scode{connect}.

\para{Reconstructing Call Sites Using Pointer-Aware Slicing} After detecting
call sites, we wish to augment them with additional information regarding the
source of each argument. This is useful because the source of each argument
can provide a hint for the possible values that this argument can store. Moreover, as
discussed above, these are essential to allow name prediction in the face of
API name obfuscation.

\ignore{
	\uri{You're talking about the "the process of creating our augmented call sites-based representation" as if the reader knows what is it. There's a first-sentence missing, saying something like: "After detecting call sites, we wish to augment them with additional information regarding the source of each argument. This is useful because the source of each argument can hint for the possible values that this argument can store. In obfuscated API settings, this is especially useful, because..." and then you can bring the paragraph that discusses obfuscation from the previous page.}
}

\begin{figure*}[]
\includegraphics[keepaspectratio, width=\textwidth]{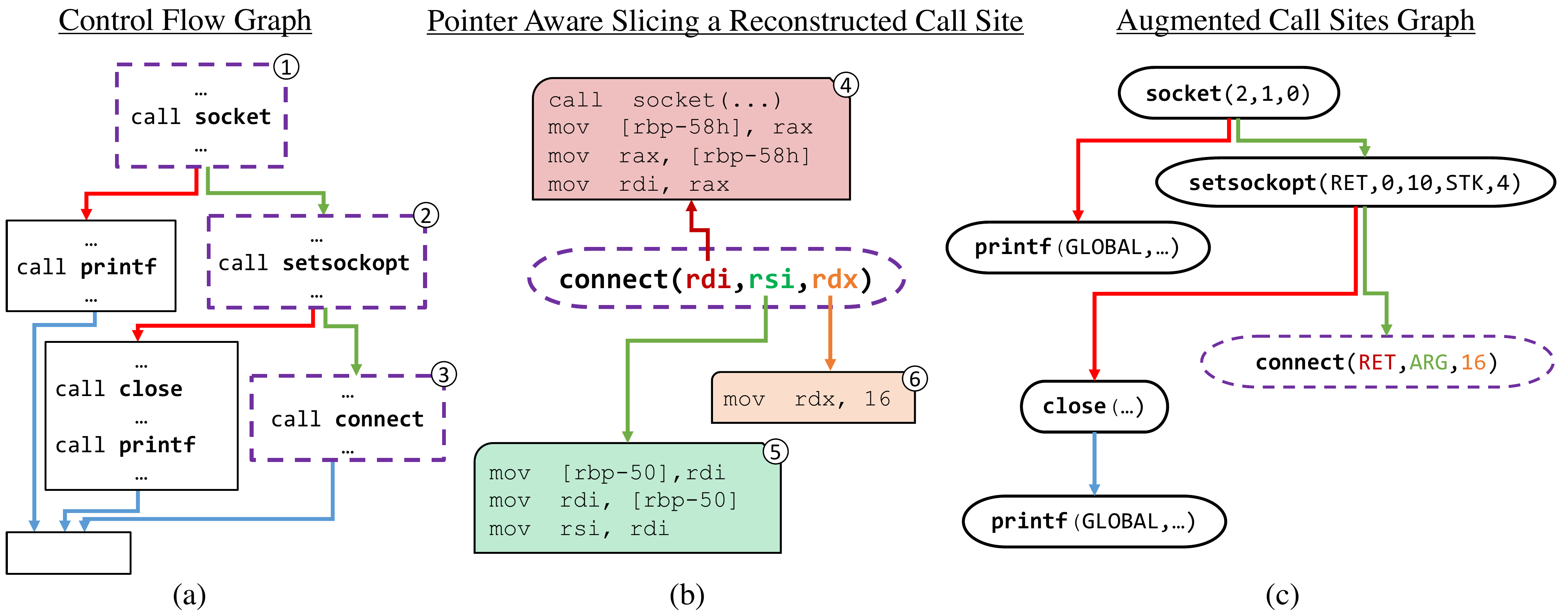}
\caption{(a) The procedure's \ac{CFG} showing all call instructions, and focusing on the path leading to the \scode{call connect} instruction is shown in dashed purple; (b) The reconstructed call site for \scode{connect} connecting each argument register to its \emph{pointer aware slice} ; (c) the augmented call sites graph containing the augmented call sites for all API calls, including \scode{connect}. }
\label{Fig@Overview@ACSG}
\end{figure*}

\Cref{Fig@Overview@ACSG} depicts the process of creating our augmented call
sites-based representation, focusing on the creation of the call site for the
\scode{connect} \ac{API} call. \Cref{Fig@Overview@ACSG}(a) depicts $P$'s
\ac{CFG} showing only call instructions.

We enrich the API call into a structure more similar to a call site in
higher-level languages. To do so, we retrieve the number of arguments passed
when calling an \ac{API} by examining debug symbols for the library from
which they were imported. Following the \ac{OS} calling convention, we map
each argument to the register used to pass it. 
By retrieving the three
arguments for the \scode{connect} API, we can reconstruct its
call site: \scode{connect(rdi, rsi, rdx)}, as shown in
\cref{Fig@Overview@ACSG}(b). This reconstructed call site contains the API
name (\scode{connect}) and the registers that are used as arguments for this API call
(\scode{rdi, rsi, rdx}).

To obtain additional information from these call sites, we examine how the
\emph{value} of each argument was computed. To gather this information, we
extract a static \emph{slice} of each register at the call location in the
procedure. A slice of a program \cite{SLICING} at a specific location for a
specific value is a subset of the instructions in the program that is necessary to
create the value at this specific location.

\textcircled{\raisebox{0pt}{\footnotesize{1}}},
\textcircled{\raisebox{0pt}{\footnotesize{2}}} and
\textcircled{\raisebox{0pt}{\footnotesize{3}}} in \cref{Fig@Overview@ACSG}(a)
mark the \textcolor{ourpurple}{purple} nodes composing the \ac{CFG} path leading to the \scode{call
connect} instruction. In \cref{Fig@Overview@ACSG}(b), each register of the
\scode{connect} call site is connected by an arrow to the slice of $P$ used to compute its value. These slices are extracted from instructions in
this path. As some arguments are pointers, we perform a \emph{pointer-aware}
static program slice. This process is explained in \cref{Sec@Repr}. 

\para{Augmenting call sites using concrete and abstract values}
Several informative slices are shown \cref{Fig@Overview@ACSG}(b):

\begin{enumerate} 

\item In \textcircled{\raisebox{0pt}{\footnotesize{4}}}, \scode{rdi} is assigned with the
return value of the previous call to \scode{socket}: 
\begin{inparaenum}[(i)] 
\item in accordance with the calling conventions, the return value of \scode{socket} is
placed in \scode{rax} before returning from the call, 
\item \scode{rax}'s value is assigned to a local variable on the stack at location \scode{rbp-58h}, 
\item \scode{rax} reads this value from the same stack
location \footnote{These seemingly redundant computation steps, \eg placing a value
on the stack and then reading it back to the same register, are necessary to
comply with calling conventions and to handle register overflow at the
procedure scope.} and finally, 
\item this value is assigned from \scode{rax} to \scode{rdi}. \end{inparaenum}

\item In \textcircled{\raisebox{0pt}{\footnotesize{5}}} , \scode{rsi} gets its
value from an argument passed into $P$: \begin{inparaenum}[(i)] \item
\scode{rdi} is coincidently used to pass this argument to $P$, \item this
value is placed in a stack variable at \scode{rbp-50h}, \item this value is
assigned from the stack into \scode{rdi}, and finally \item this value is
assigned from \scode{rdi} to \scode{rsi} \end{inparaenum}. 

\item In \textcircled{\raisebox{0pt}{\scriptsize{6}}}, the constant value of
$16$ is assigned directly to \scode{rdx}. This means that we can
\emph{determine the concrete value} for this register that is used as an argument of
\scode{connect}. 

\end{enumerate} 

Note that not all the instructions in \cref{Fig@Overview@ACSG}(b)'s slices
appear above the call instruction in \cref{Fig@ASM@Dissassmby}. This is caused
by the compiler optimizations and other placements constraints. \TODO{\uri{To difficult to follow. I didn't understand the problem and didn't understand the reason. If it's not critical, maybe we can remove this note.}}

In slices \textcircled{\raisebox{0pt}{\footnotesize{4}}} and \textcircled{\raisebox{0pt}{\footnotesize{5}}}, the concrete values
for the registers that are used as arguments \emph{are unknown}.  Using static analysis,
we \emph{augment} the reconstructed call sites by replacing the register names
(which carry no meaning) with: (i) the concrete value if such can be
determined; or (ii) a broader category which we call argument \emph{abstract
values}. An abstract value is extracted by classifying the slice into one of
the following classes: argument (\scode{ARG}), global value (\scode{GLOBAL}),
unknown return value from calling a procedure (\scode{RET}) and local variable
stored on the stack (\scode{STK}). Complete details regarding this
representation are presented in \cref{Sec@Repr}.

Performing this augmentation process on the \scode{connect} call site results
in \scode{connect(RET,ARG,16)}, as shown in \cref{Fig@Overview@ACSG}(c).
\cref{SSec@Eval@Ablation} of our evaluation shows that augmenting call sites
using abstract and concrete values provides our model a relative improvement of 
 $\NeroAugmentImprove$ over the alternative of using only the API name.
Further, as we show in \cref{SSec@Eval@Debin}, abstract and concrete values
allow the model to make predictions even when API names are obfuscated. 

\para{Predicting procedure names using augmented call site graph}
\Cref{Fig@Overview@ACSG}(c) shows how augmented call sites are used to create
the augmented call sites graph. Each node represents an augmented call site.
Call site nodes are connected by edges representing possible run-time
sequences of $P$. This depicts how \emph{correctly ordered augmented call
sites} create a powerful representation for binary procedures. This becomes
especially clear compared to the disassembled code
(\cref{Fig@ASM@Dissassmby}).

Using this representation, we can train a \ac{GNN} based model on the graph
itself. Alternatively, we can employ \ac{LSTM}-based and Transformer-based models by extracting
simple paths from the graph, to serve as sequences that can be fed into these models.

\ignore{
Focusing on the sequence of call sites shown in \cref{Fig@ASM@CallSequence}
with all their abstract values, we observe that:  \begin{inparaenum}[(i)]	
\item \scode{memset(\tiny{STK,0,48}\small{)}} initializes a 48-byte memory
space with zeroes;   \item \scode{getaddrinfo(\tiny{ARG,ARG,STK,STK}\small{)}}
uses two of $P's$ arguments to search for a specific interface to be used
later;  \item \scode{socket(\tiny{STK,STK,STK}\small{)}} and
\scode{setsocketopt(\tiny{RET,1,2,STK,4}\small{)}} create a socket and
configure it to be a TCP socket by passing the value \scode{1};  and \item
\scode{bind} and \scode{listen} determine that this procedure is part of a
server listening to incoming connections.   \end{inparaenum} 
}
 
\para{Key aspects} The illustrated example highlights several key aspects of our approach:
\begin{itemize}
    \item Using static analysis, augmented API call sites can be reconstructed from the shallow assembly code.
    \item Analyzing pointer-aware slices for arguments allows call site augmentation by replacing register names with concrete or abstracted values.
    \item By analyzing the \ac{CFG}, we can learn augmented call sites in their approximated runtime order.
    \item A \ac{CFG} of \emph{augmented call site} is an efficient and informative representation of a binary procedure.
    \item Various neural network architectures trained using this representation can accurately predict complex procedure names.
\end{itemize}

%% file: background.tex
\section{Background}\label{Sec@Background} 

In this section we provide the necessary background for the neural models that we later use to demonstrate our approach. In
\cref{SSec@Background@encoderdecoder}  we describe the encoder-decoder
paradigm which is the basis for most seq2seq models; in
\cref{SSec@Background@attention} we describe the mechanism of attention;
in \cref{SSec@Background@transformers} we describe Transformer models; and in \cref{SSec@Background@gnns} we describe graph neural networks (GNNs).

\para{Preliminaries} Contemporary seq2seq models are usually based on the encoder-decoder paradigm
\cite{cho2014learning, sutskever2014sequence}, where the encoder maps a
sequence of input symbols $\boldsymbol{x}=\left(x_1,...,x_n\right)$ to a
sequence of latent vector representations
$\boldsymbol{z}=\left(z_1,...,z_n\right)$. Given $\boldsymbol{z}$, the decoder
predicts a sequence of target symbols
$\boldsymbol{y}=\left(y_1,...,y_m\right)$, thus modeling the conditional
probability: $p\left(y_1,...,y_m\mid x_1,...,x_n\right)$. At each decoding time
step, the model predicts the next symbol conditioned on the previously
predicted symbol, hence the probability of the target sequence is
factorized as:
\begin{equation*}
p\left(y_1,...,y_m\mid x_1,...,x_n\right)=\prod_{t=1}^{m}p\left(y_t\mid y_{<t},z_1,...,z_n\right)
\end{equation*}

\subsection{LSTM Encoder-Decoder Models} \label{SSec@Background@encoderdecoder}

Traditionally \cite{cho2014learning, sutskever2014sequence}, seq2seq models
are based on \acp{RNN}, and typically \acs{LSTM}s \cite{hochreiter1997lstm}. LSTMs are trainable neural network components that work on a sequence of
\emph{input vectors}, and return a sequence of output vectors, based on
internal learned weight matrices. Throughout the processing of the input sequence,
the \ac{LSTM} keeps an internal \emph{state} vector that is updated after
reading each input vector.

The encoder embeds each input symbol into a vector using a learned embedding matrix $E^{in}$.
The encoder then maps the input symbol embeddings to a sequence
of latent vector representations $\left(z_1,...,z_n\right)$ using an encoder LSTM:
\begin{align*}
z_1,...,z_n &= \text{LSTM}_{encoder}\left(\text{embed}\left(E^{in}, x_1,...,x_n\right)\right) 
\end{align*}
The decoder is an additional LSTM with separate learned weights. The decoder uses 
an aggregation of the encoder LSTM states as its initial state; traditionally,  the final hidden state of the encoder LSTM is used:
\begin{align*}
h^{dec}_1,...,h^{dec}_m &= \text{LSTM}_{decoder}\left(z_n\right)
\end{align*}

At each decoding step, the decoder reads a target symbol and
outputs its own state vector $h^{dec}_t$ given the currently fed target symbol
and its previous state vector. The decoder then computes a dot product between its new
state vector $h^{dec}_{t}$ and a learned embedding
vector $E^{out}_i$ for each
possible output symbol in the vocabulary $y_i \in Y$, and normalizes
the resulting scores to get a distribution over all possible symbols:
\begin{align}
&p\left(y_{t}\mid y_{<t},z_1,...,z_n\right)
=\text{softmax}\left(E^{out}\cdot h^{dec}_t\right) %
\label{Eq@Model@encoderdecoder@noattentionprob}
\end{align}

where $\text{softmax}$ is a function that takes a vector of scalars and normalizes it into a probability distribution.
That is, each dot product $E_i^{out}\cdot h^{dec}_t$ produces a scalar score
for the output symbol $y_i$, and these scores are normalized by exponentiation
and division by their sum. This results in a probability distribution over the
output symbols $y\in Y$ at time step $t$.

The target symbol that the decoder reads at time step $t$ differs between training and test time: at test time, the decoder reads the estimated target symbol that the decoder itself has predicted in the previous step $\hat{y}_{t-1}$; at training time, the decoder reads the ground truth target symbol of the previous step $y_{t-1}$.
This training setting is known as ``teacher forcing'' - even if the decoder makes an incorrect prediction at time $t-1$, the true $y_{t-1}$ will be used to predict $y_t$. At test time, the information of the true $y_{t-1}$ is unknown.

\subsection{Attention Models}\label{SSec@Background@attention}

In attention-based models, at each step the decoder has the ability to compute a different weighted average of \emph{all} latent vectors $\boldsymbol{z}=\left(z_1,...,z_n\right)$ \cite{luong15, bahdanau14}, and not only the last state of the encoder as in traditional seq2seq models \cite{sutskever2014sequence, cho2014learning}. The weight that each $z_i$ gets in this weighted average can be thought of as the \emph{attention} that this input symbol $x_i$ is given at a certain time step. This weighted average is produced by computing a score for each $z_i$ conditioned on the current decoder state $h_t^{dec}$. These scores are normalized such that they sum to $1$:
\begin{align*}
\boldsymbol{\alpha}_t=\text{softmax}\left(\boldsymbol{z} \cdot W_a \cdot h^{dec}_t\right)
\end{align*}

That is, $\boldsymbol{\alpha}_t$ is a vector of positive numbers that sum to
$1$, where every element ${{\alpha_t}_i}$ in the vector is the normalized score
for $z_i$ at decoding step $t$. $W_a$ is a learned weights matrix that
projects $h^{dec}_t$ to the same size as each $z_i$, such that dot product can
be performed.

Each $z_i$ is multiplied by its normalized score to produce $c_t$, the context
vector for decoding step $t$:
\begin{align*}
c_t = \sum_{i}^{n}{\alpha_t}_i \cdot z_i
\end{align*}

That is, $c_t$ is a weighted average of
$\boldsymbol{z}=\left(z_1,...,z_n\right)$, such that the weights are
conditioned on the current decoding state vector $h^{dec}_t$. The dynamic
weights $\boldsymbol{\alpha}_t$ can be thought of as the \emph{attention} that
the model has given to each $z_i$ vector at decoding step $t$.

The context vector $c_t$ and the decoding state $h^{dec}_t$ are then combined
to predict the next target token $y_t$. A standard approach \cite{luong15} is
to concatenate $c_t$ and $h^{dec}_t$ and pass them through another learned linear layer $W_c$ and a nonlinearity $\sigma$, to predict the next symbol:
\begin{align}
\widetilde{h_t^{dec}}=&\sigma\left(W_c \left[c_t ; h^{dec}_t\right]\right) \nonumber\\
p\left(y_{t}\mid y_{<t},z_1,...,z_n\right) = &\text{softmax}\left(E^{out}\cdot \widetilde{h_t^{dec}}\right)
\label{Eq@Model@encoderdecoder@attentionprob}
\end{align}

Note that target symbol prediction in non-attentive models
(\cref{Eq@Model@encoderdecoder@noattentionprob}) is very similar to the
prediction in attention models (\cref{Eq@Model@encoderdecoder@attentionprob}),
except that non-attentive models use the decoder's state $h_t^{dec}$ to
predict the output symbol $y_t$, while attention models use
$\widetilde{h_t^{dec}}$, which is the decoder's state \emph{combined} with the
dynamic context vector $c_t$. This dynamic context vector $c_t$ allows the
decoder to focus its attention to the most relevant input vectors at different
decoding steps.

\subsection{Transformers}\label{SSec@Background@transformers}
Transformer models were introduced by \citet{vaswani2017attention} and were shown to outperform LSTM models for many seq2seq tasks. Transformer models are the basis for the vast majority of recent sequential models \cite{devlin2019bert, radford2019language, chiu2018state}.

The main idea in Transformers is to rely on the mechanism of self-attention without any recurrent components such as LSTMs.
Instead of reading the input vectors $\boldsymbol{x}=\left(x_1,...,x_n\right)$ \emph{sequentially} as in \acs{RNN}s, the input vectors are passed through several layers, which include multi-head \emph{self-}attention and a fully-connected feed-forward network. 

In a self-attention layer, each $x_i$ is mapped to a new value that is based on attending to all other $x_j$'s in the sequence. First, each $x_i$ in the sequence is projected to query, key, and value vectors using learned linear layers:
\begin{align*}
Q = W_{q}\cdot \boldsymbol{x} & & K = W_{k}\cdot \boldsymbol{x} & & V = W_{v} \cdot \boldsymbol{x} 
\end{align*}
where $Q\in \mathbb{R}^{n \times d_k}$ holds the queries, $K\in \mathbb{R}^{n \times d_k}$ holds the keys, and $V\in \mathbb{R}^{n \times d_v}$ holds the values. The output of self-attention for each $x_i$ is a weighted average of the other value vectors, where the weight assigned to each value is computed by a compatibility function of the query vector of the element $Q_i$ with each of the key vectors of the other elements $K_j$, scaled by $\frac{1}{d_k}$. This can be performed for all keys and queries in parallel using the following quadratic computation:
\begin{align*}
\alpha\left(Q,K\right)=\text{softmax}\left(\frac{QK^\top}{\sqrt{d_k}}\right)
\end{align*}

That is, $\alpha \in \mathbb{R}^{n \times n}$ is a matrix in which every entry $\alpha_{ij}$ contains the scaled score between the query of $x_i$ and the key of $x_j$.
These scores are then used as the factors in the weighted average of the value vectors: 
\begin{align*}
\text{Attention}\left(Q,K,V\right)=\alpha\left(Q,K\right) \cdot V
\end{align*}
In fact, there are multiple separate (typically 8 or 16) attention mechanisms, or ``heads'' that are computed in parallel. Their resulting vectors are concatenated and passed through another linear layer. There are also multiple (six in the original Transformer) layers, which contain separate multi-head self-attention components, stacked on top of each other.

The decoder has similar layers, except that one self-attention mechanism attends to the previously-predicted symbols (i.e., $y_{<t}$) and another self-attention mechanism attends to the encoder outputs (i.e., $z_1,...,z_n$).

The Transformer model has a few more important components such as residual connections, layer normalization and positional embeddings, see \citet{vaswani2017attention} and excellent guides such as ``The Illustrated Transformer'' \cite{illustratedTransformer}.

\subsection{Graph Neural Networks}
\label{SSec@Background@gnns}

A directed graph $\mathcal{G}=\left(\mathcal{V},\mathcal{E}\right)$ contains nodes $\mathcal{V}$ and edges $\mathcal{E}$, where $\left(u,v\right)\in\mathcal{E}$ denotes an edge from a node $u$ to a node $v$, also denoted as $u\rightarrow v$. For brevity, in the following definitions we treat all edges as having the same \emph{type}; in general, every edge can have a type from a finite set of types, and every type has different learned parameters.

Graph neural networks operate by propagating neural messages between neighboring nodes. At every propagation step, the network computes each node's sent messages given the node's current representation, and every node updates its representation by aggregating its received messages with its previous representation. 

Formally, at the first layer ($k=0$) each node is associated with an initial representation $\mathbf{h}_v^{\left(0\right)} \in \mathcal{R}^{d_0}$. This representation is usually derived from the node's label or its given features. Then, a GNN layer updates each node's representation given its neighbors, yielding $\mathbf{h}_v^{\left(1\right)} \in \mathcal{R}^{d}$. In general, the $k$-'th layer of a GNN is a parametric function $f_k$ that is applied independently to each node by considering the node's previous representation and its neighbors' representations:
\begin{equation}
	\mathbf{h}_v^{\left(k\right)}=f_k\left(
	\mathbf{h}_v^{\left(k-1\right)}, 
	\{\mathbf{h}_u^{\left(k-1\right)}\mid u\in\mathcal{N}_v\}
	; \theta_{k}\right)
	\label{eq:layer}
\end{equation}
Where $\mathcal{N}_v$ is the set of nodes that have  edges to $v$: $\mathcal{N}_v=\{u \in \mathcal{V} \mid \left( u,v \right) \in \mathcal{E}\}$. 
The same $f_k$ layer weights can be unrolled through time;
alternatively, each $f_k$ can have weights of its own, increasing the model's capacity by using different weights for each value of $k$. The total number of layers $K$ is usually determined as a hyperparameter. 

The design of the function $f$ is what mostly distinguishes one type of GNN from the other. For example, graph convolutional networks \cite{kipf2017semi} define $f$ as:
\begin{equation}
	\mathbf{h}_v^{\left(k\right)}=
	\sigma\left( \sum_{u\in \mathcal{N}_v}  \frac{1}{c_{u,v}} 
	W_{neighbor}^{\left(k\right)}\mathbf{h}_{u}^{\left({k-1}\right)} 
+ 	W_{self}^{\left(k\right)}\mathbf{h}_{v}^{\left({k-1}\right)} 
	\right)
	\label{eq:gcn}
\end{equation}
Where $\sigma$ is an activation function such as $ReLU$, and $c_{u,v}$ is a normalization factor that is often set to $|\mathcal{N}_v |$ or $\sqrt{|\mathcal{N}_v | \cdot |\mathcal{N}_u} |$. 

To generate a sequence from a graph, in this work we used an LSTM decoder that attends to the final node representations, as in \cref{SSec@Background@attention}, where the latent vectors $\left(z_1,...,z_n\right)$ are the final node representations  $\{\mathbf{h}_v^{\left(K\right)} \mid v \in \mathcal{V}\}$.

%% file: representation.tex
\section{Representing Binary Procedures for Name Prediction} \label{Sec@Repr}
The core idea in our approach is to predict procedure names from structured augmented call sites. In~\cref{Sec@Model} we will show how to employ the neural models of~\cref{Sec@Background} in our setting, but first, we describe how we produce structured augmented call sites from a given binary procedure.

In this section, we describe the program analysis that builds our
representation from a given binary procedure.

\para{Pre-processing the procedure's CFG} Given a binary procedure $P$, we
construct its \ac{CFG}, $G_P$: a directed graph composed of nodes which
correspond to P's \acp{BB}, connected by edges according to control flow
instructions, i.e., jumps between basic-blocks. An example for a \ac{CFG} is shown in \cref{Fig@Overview@ACSG}(b). For clarity, in the following:  \begin{inparaenum}[(i)] 
\item we add an artificial entry node \ac{BB} denoted $Entry$, and connect it to the original entry \ac{BB} and, 
\item we connect all exit-nodes, in our case any \ac{BB} ending with a return instruction (exiting the procedure), to another artificial sink node $Sink$.
\end{inparaenum}

For every simple path $p$ from $Entry$ to $Sink$ in $P_G$, we use
$instructions(p)$ to denote the sequence of instructions in the path \acp{BB}.
We use $\semp{P}$ to denote the set of sequences of instructions along simple
paths from $Entry$ to $Sink$, that is: \[ \semp{P} = \{ instructions(p) \mid p
\in simplePaths(Entry,Sink) \}.\]

Note that when simple paths are extracted, loops create (at least) two paths:
one when the loop is not entered, and another where the loop is unrolled once. 

From each sequence of instructions $s \in \semp{P}$, we extract a list of
\scode{call} instructions. Call instructions contain the call target, which is
the address of the procedure called by the instruction, henceforth target
procedure. 
There are three options for call targets:
\begin{itemize} 
\item Internal: an internal fixed address of a procedure inside the executable.
\item External: a procedure in external dynamically loaded code. 
\item Indirect: a call to an address stored in a register at the time of the call. (\eg call
rax). \end{itemize}

\para{Reconstructing call sites} In this step of our analysis, we will
reconstruct the call instructions, \eg \scode{call 0x40AABBDD}, into a
structured representation, \eg \scode{setsockopt($arg_1$, ..., $arg_k$)}. This
representation is inspired by call sites used in higher programming languages
such as C. This requires resolving the name of the target procedure and the
number of arguments sent ($k$ in our example).

For internal calls, the target procedure name is not required during runtime,
and thus it is removed when an executable is stripped. The target procedure name of
these calls cannot be recovered. Given a calling convention, the number of
arguments can be computed using data-flow analysis. Note that our analysis is
static and works at the intra-procedure level, so a recursive call is handled
like any other internal call. 

Reconstructing external calls requires the information about external
dependencies stored in the executable. Linux executables declare external
dependencies in a special section in the \ac{ELF} header. This causes the
\ac{OS} loader to load them dynamically at runtime and allows the executable
to call into their procedures. These procedures are also called imported
procedures. These external dependencies are usually software libraries , such
as ``libc'', which provides the API for most \ac{OS} related functionality
(\eg opening a file using \scode{open()}). 

The target procedure name for external calls is present in the \ac{ELF}
sections. The number of arguments can be established by combining information
from: the debug information for the imported library, the code for the
imported library, and the calling procedure code preparing these arguments for
the call. In our evaluation (\Cref{Sec@Eval}), we explore cases in which the procedure target name from the
\ac{ELF} sections is obfuscated and there is no debug information for imported
code.

In the context of a specific CFG path, some indirect calls can be resolved into
a an internal or external call and handled accordingly. For unresolved
indirect calls, the number of arguments and target procedure name remain
unknown. 

Using the information regarding the name and number of arguments for the
target procedure (if the information is available), we reconstruct the call
site according to the \ac{ABI}, which dictates the calling convention. In our
case (System-V-AMD64-ABI) the first six procedure arguments that fit in the
native 64-bit registers are passed in the following order: \scode{rdi, rsi,
rdx, rcx, r8, r9}. Return values that fit in the native register are returned
using \scode{rax}. For example, libc's \scode{open()} is defined by the
following C prototype: \scode{int open(const char *pathname, int flags,
mode\_t mode)}. As all arguments are pointers or integers, they can be passed
using native registers, and the reconstruction process results in the
following call site: \scode{rax = open(rdi, rsi, rdx)}.

The Linux \ac{ABI} dictates that all global procedures (callable outside the
compilation unit yet still internal) must adhere to the calling convention.
Moreover, a compiler has to prove that the function is only used inside the
compilation unit, and that its pointer does not ``escape'' the compilation unit
to perform any \ac{ABI} breaking optimizations. We direct the reader to the
Linux ABI reference (\citeauthor{LINUXABI}) for more information about calling
conventions and for further details on how floats and arguments exceeding the
size of the native registers are handled. 

\ignore{
\input{tables/reconstructions.tex}

\Cref{TAB@Eval@ReconstructTAB} summarizes this last section.
}

\subsection{Augmenting call sites with concrete and abstract values}

Our reconstructed call site contains more information than a simple call
instruction, but is still a long way from a call site in higher level programming
language. As registers can be used for many purposes, by being assigned a
specific value at different parts of the execution, their presence in a call
site carries little information. To augment our call site based representation
our aim is to replace registers with a concrete value or an abstract value. Concrete
values, \eg ``1'', are more informative than abstract values, yet a suitable
abstraction, \eg ``RET'' for a return value, is still better than a register
name.

\para{Creating pointer-aware slice-trees} To ascertain a value or abstraction
for a register that is used as an argument in a call, we create a pointer-aware slice
for each register value at the call site location. This slice contains the
calculation steps performed to generate the register value. We adapt the
definitions of \citet{PTR_SLICING} to assembly instructions. For an
instruction $inst$ we define the following sets:

\begin{itemize}
\item $V_{write}(inst)$: all registers that are written into in $i$.
\item $V_{read}(inst)$: all values used or registers read from in $i$.
\item $P_{write}(inst)$: all memory addresses written to in $i$.
\item $P_{read}(inst)$: all memory addresses read from in $i$.
\end{itemize}

The first two sets, $V_{read}(inst)$ and $V_{write}(inst)$, capture the
\emph{values} that are read in the specific instruction and the values that
are written into in the instruction. The last two sets, $P_{read}(inst)$ and
$P_{write}(inst)$, capture the \emph{pointers} addressed for writing and
reading operations. Note that the effects of all control-flow instructions on
the stack and the \scode{EIP} and \scode{ESP} registers are excluded from
these definitions (as they are not relevant to data-flow).

\input{code/BuildingSliceInfo.tex}

\Cref{Tab@Repr@Sliceing} shows three examples for assembly instructions and
their slice information sets.

Instruction $inst_1$ shows a simple assignment of a constant value into the
\scode{rax} register. Accordingly, $V_{read}(inst_1)$ contains the value
\scode{5} of the constant that is read in the assignment. $V_{write}(inst_1)$
contains the assignment's target register \scode{rax}. 

Instruction $inst_2$ shows an assignment into the same register, but this time,
from the memory value stored at offset \scode{rbx+5}. The full
expression \scode{rbx+5} that is used to compute the memory address to read
from is put in $P_{read}$, and the register used to compute this address is
put in $V_{read}$. Note that $V_{read}(inst_1)$ contains the value \scode{5}, 
because it is used as a constant value, while $V_{read}(inst_2)$ does not contain the
value $5$ because in $inst_2$ it is used as a memory offset.

Instruction $inst_3$ shows an indirect call instruction. This time, as the callee
address resides in \scode{rcx}, it is put in $V_{read}$. $V_{write}(inst_3)$
contains $rax$, which does not explicitly appear in the assembly instruction,
because the calling convention states that return values are placed in $rax$.

Given an instruction sequence $s \in \semp{P}$, we can define:

\begin{itemize}
\item $V_{last-write}(inst,r)$: the maximal instruction $inst'<inst$, in which for the register $r$:  $r \in V_{write}(inst')$.
\item $P_{last-write}(inst,p)$: the maximal instruction $inst'<inst$, in which for the pointer $p$: $p \in P_{write}(inst')$.
\end{itemize}

\begin{figure*}[h]
\includegraphics[keepaspectratio, width=\textwidth]{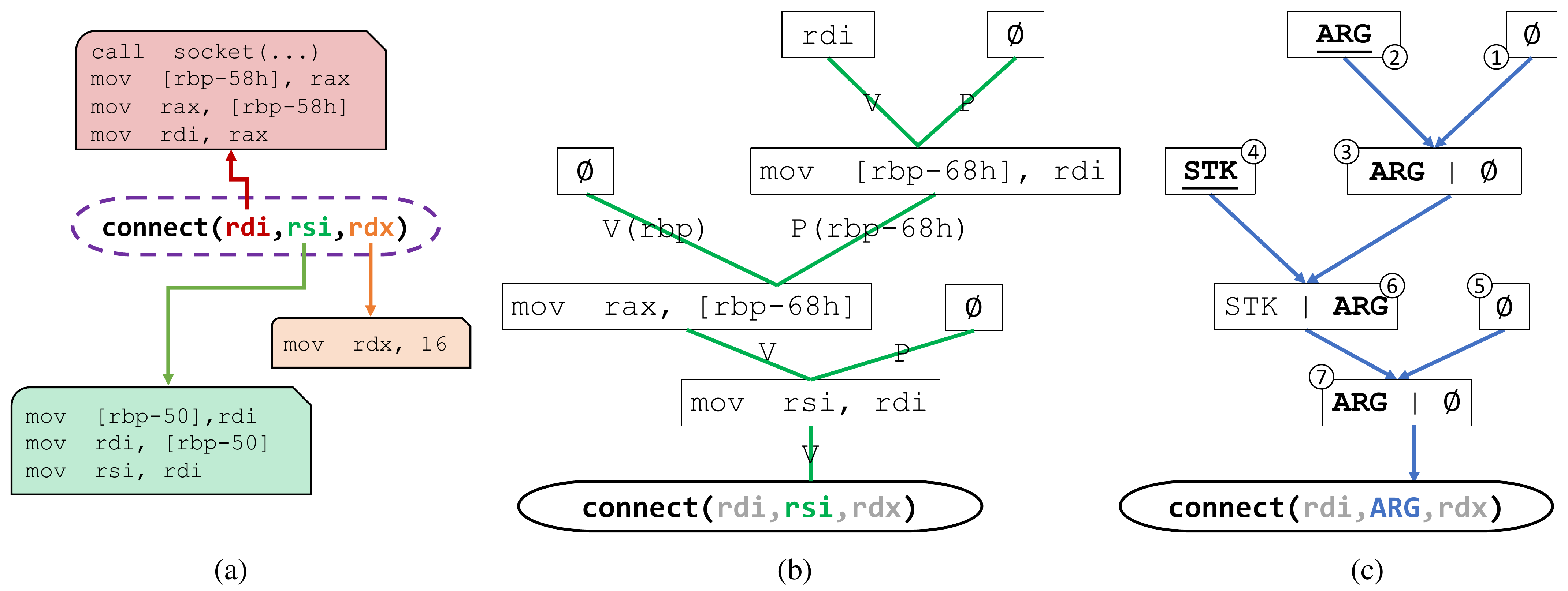}
\caption{Creating pointer-aware slices for extracting concrete values or abstract values: all possible computations that create the value of the \scode{rsi} are statically analyzed, and finally, the occurrence of \scode{rsi} in the call to \scode{connect} is replaced with the ARG \emph{abstract value} originating from \textcircled{\raisebox{0pt}{\footnotesize{2}}}, which is defined as more informative than $\emptyset$ (\textcircled{\raisebox{0pt}{\footnotesize{1}}}) and STK (\textcircled{\raisebox{0pt}{\footnotesize{4}}}).}
\label{Fig@Repr@SliceTrees}
\end{figure*}

Finally, to generate a pointer-aware slice for a specific register $r$ at a
call site in an instruction $inst$, we repeat the following calculations:
$V_{last-write} \rightarrow V_{read}$ and $P_{last-write} \rightarrow
P_{read}$, until we reach empty sets. Although this slice represents a
sequential computation performed in the code, it is better visualized as a
tree. \cref{Fig@Repr@SliceTrees}(a) shows the pointer-aware slice for the
\scode{rsi} register (\textcolor{ourgreen}{green}) that is used as an argument for \scode{connect};
\cref{Fig@Repr@SliceTrees}(b) represents the same slice as a tree, we denote as a ``slice tree''.

\para{Analyzing slice trees}  We analyze slice trees to select a concrete
value or an abstract value to replace the register name in the \emph{sink} of the
tree.\footnote{Formally, the root of the tree}  To do this, we need to assign
abstract values to some of the slice tree leaves, and propagate the values in
the tree.

We begin by performing the following initial tag assignments in the tree: \begin{itemize}
\item The arguments accepted by $P$ receive the ``ARG'' tag.

\item All \scode{call} instructions receive with the ``RET'' tag. Note that
this tag will allways be propagated to a \scode{rax} register as it holds the
return value of the called procedure.

\item The \scode{rbp} register at the beginning of $P$ receives the ``STK''
tag. During the propagation process, this tag might be passed to \scode{rsp}
or other registers.\footnote{While it was a convention to keep \scode{rbp}
and \scode{rsp} set to positions on the stack -- modern compilers often use
these registers for other purposes.} 

\item When a pointer to a global value can not be resolved to a concrete
value, \eg a number or a string, this pointer receive the ``GLOBAL''
tag.\footnote{One example for such case is a pointer to a global structure.}

\end{itemize}

After this initial tag assignment, we propagate values downwards in the tree
towards the register at the tree sink. We prefer values according to the
following order: (1) concrete value, (2) ARG, (3) GLOBAL, (4) RET, (5) STK.
This hierarchy represents the certainty level we have regarding a specific
value: if the value was created by another procedure -- there is no
information about it; if the value was created using one of $P$'s arguments, we
have some information about it;  and if the concrete value is available -- it
is used instead of an abstract value. To make this order complete, we add (6)
$\emptyset$ (empty set), but this tag is never propagated forward over any
other tag, and in no case will reach the sink. This ordering is required to
resolve cases in which two tags are propagated into the same node. An example
for such a case is shown in the following example.

\Cref{Fig@Repr@SliceTrees}(c) shows an example of this propagation
process:

\begin{enumerate}

\item The assigned tags ``ARG'' and ``STK'' in
\textcircled{\raisebox{0pt}{\footnotesize{2}}} and
\textcircled{\raisebox{0pt}{\footnotesize{4}}}, accordingly, are marked in bold
and underlined.

\item Two tags, ``ARG'' and ``$\emptyset$'' are propagated to
\textcircled{\raisebox{0pt}{\footnotesize{3}}}, and ``ARG'', marked in bold,
is propagated to \textcircled{\raisebox{0pt}{\footnotesize{6}}}.

\item In \textcircled{\raisebox{0pt}{\footnotesize{6}}}, as ``ARG'' is
preferred over ``STK'', ``ARG'', marked again in bold, is propagated to
\textcircled{\raisebox{0pt}{\footnotesize{7}}}.

\item From \textcircled{\raisebox{0pt}{\footnotesize{6}}} ``ARG'', in bold, is
finally propagated to the sink and thus selected to replace the \scode{rdi}
register name in the \scode{connect} call site.

\end{enumerate}

\begin{figure*}[h]
\includegraphics[keepaspectratio, width=\textwidth]{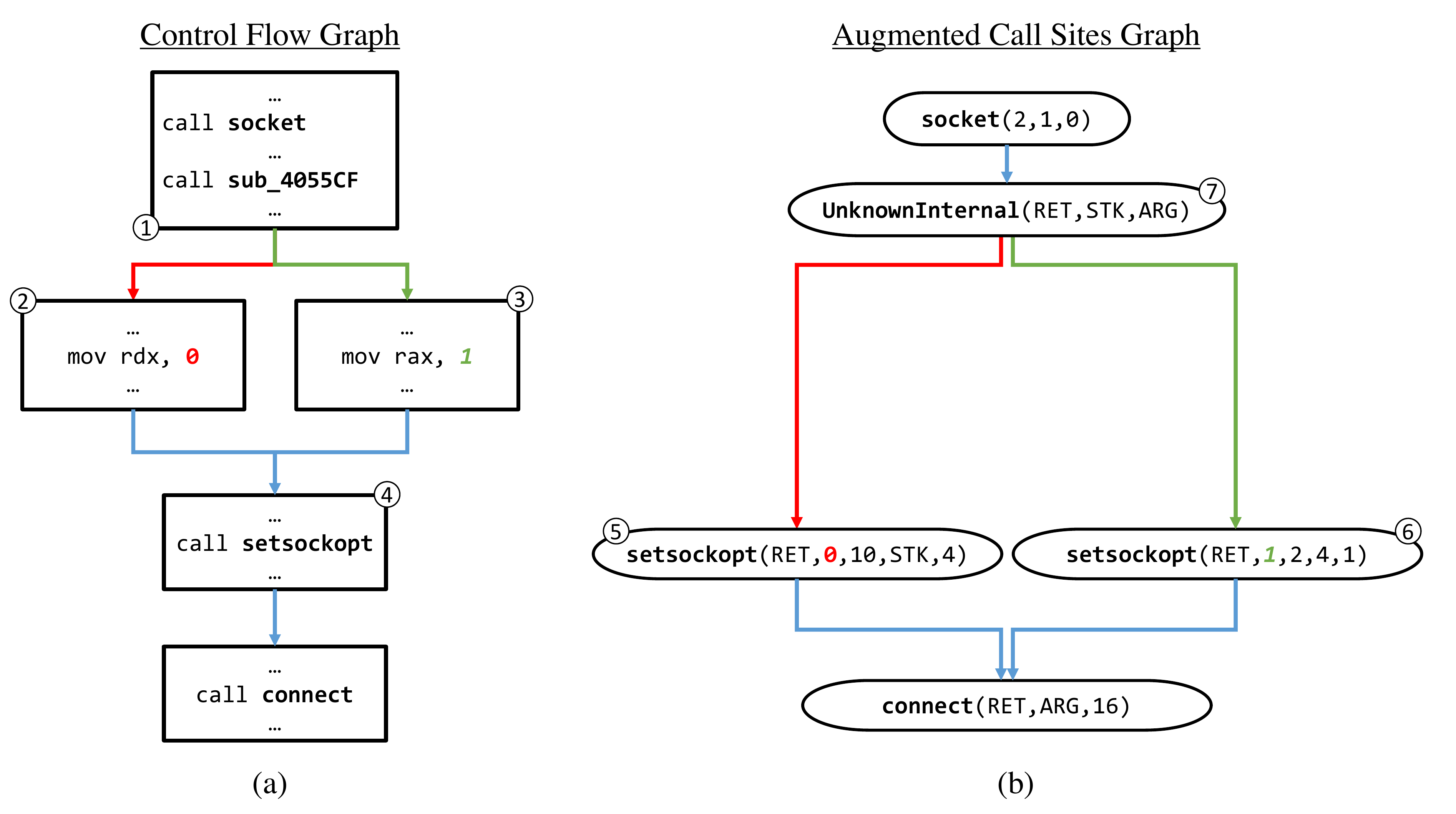}
\caption{An example of a \ac{CFG} and the augmented call site graph created by analyzing it using our approach}
\label{Fig@Repr@ACSG_ext}
\end{figure*}

\pagebreak

\para{An augmented call site graph example} \cref{Fig@Repr@ACSG_ext} shows an
example of a \ac{CFG} and the augmented call site graph, henceforth augmented
graph for short, created by analyzing it using our approach.

Node \cbox{4} from the \ac{CFG} is represented by two nodes in the
augmented graph: \cbox{5} and \cbox{6}. This is the result of two different
paths in the \ac{CFG} leading into the call instruction \scode{call
setsockopt}: \cbox{1}$\rightarrow$\cbox{2}$\rightarrow$\cbox{4} and
\cbox{1}$\rightarrow$\cbox{3}$\rightarrow$\cbox{4}. In each path, the values building the augmented call site are different due to the different slices created for \emph{the same registers used as arguments} in each path. One such difference is the second argument of \scode{setsockopt} being a ({\color{red}{red}}) {\color{red}{$0$}} in \cbox{5} and a ({\color{ourgreen}{green}}) {\color{ourgreen}{$1$}} in \cbox{6}. 

Also note that internal procedure calls, while missing the target procedure name,
which is marked by ``UnknownInternal'' in \cbox{7}, is part of the augmented
graph and thus contribute to the representation of $P$.

\para{Dealing with aliasing and complex constant expressions} Building upon
our previous work, \citet{GITZ}, we re-optimize the entire procedure and
re-optimize every slice before analyzing it to deduce argument values. A
complex calculation of constants is thus replaced by the re-optimization
process. For example, the instructions \scode{xor rax,rax; inc rax;} are
simply replaced to \scode{mov rax,1}. This saves us from performing
the costly process of expression simplification and the need to expand our
analysis to support them. 

Furthermore, although very common in (human written) source code -- memory
aliasing is seldom performed in optimized binary code, as it contradict the
compiler's aspiration for reducing memory consumption and avoiding redundant
computations. An example for how aliasing is avoided is depicted in the lower left
(\textcolor{ourgreen}{green}) slice in \cref{Fig@Repr@SliceTrees}(a): an offset to the stack storing
a local variable can always be addressed in the same way using \scode{rbp} (in
this case, \scode{rbp-50h}). The re-optimization thus frees us from expanding
our analysis to address memory aliasing.

\para{Facilitating augmented call site graphs creation}  To efficiently
analyze  procedures with a sizable \acp{CFG} (\ie a large number of basic
blocks and edges) we apply a few optimizations: \begin{inparaenum}[(i)] \item
cache value computation for path prefixes; \item merge sub-paths if they do
not contain calls; \item remove duplicate sequences; and \item as all call instructions are replaced with the ``RET'' label, we do not continue the slicing operation after encountering call commands. \end{inparaenum}

%% file: tables/reconstructions.tex
\begin{table}[]
\begin{tabular}{@{}lll@{}}
\toprule
Procedure Type \\ Reconstructed part & Name                                             & \#Arguments                                                 \\ \midrule
Inernal                           & Unrecoverable                                    & Recovered using calling and callee code                     \\
External                          & Available in \textbackslash{}ac\{ELF\} sections* & \begin{minipage}[t]{0.8\columnwidth} Recovered using debug information**, callee and caller code \end{minipage} \\
Unresolved Indirect               & Unrecoverable                                    & Partially recoverable using caller code                     \\ \bottomrule
\end{tabular}

\caption{A summary of information used when reconstructing call sites for each call taget option.}

\label{TAB@Eval@ReconstructTAB}
\end{table}

%% file: code/BuildingSliceInfo.tex
\begin{table}[h]
\begin{tabular}{l|l|c|c|c|c}
& \multicolumn{1}{c|}{Instruction} & $V_{read}$ & $V_{write}$ & $P_{read}$ & $P_{write}$ \\ \hline
$inst_1$ & \scode{mov rax,5} & $5$ & \scode{rax} & $\emptyset$ &  $\emptyset$  \\ \hline
$inst_2$ & \scode{mov rax,[rbx+5]} & \scode{rbx} & \scode{rax} & \scode{rbx+5}  & $\emptyset$  \\ \hline
$inst_3$ & \scode{call rcx} & \scode{rcx} & \scode{rax} & \scode{rcx} &  $\emptyset$
\end{tabular}
\caption{An example for slice information sets created by three x64 instructions: $V_{read|write}$ sets show values read and written into and $P_{read|write}$ show pointer dereferences for reading from and writing to memory.}
\label{Tab@Repr@Sliceing}
\end{table}

%% file: model.tex
\section{Model}\label{Sec@Model}

The key idea in this work is to represent a binary procedure as structured call
sites, and the novel synergy between program analysis of binaries and
neural models, rather than the specific neural architecture. 
In \cref{Sec@Eval}, we experiment with plugging in our representation into LSTMs, Transformers, and GCNs.
In
\cref{SSec@Model@Callsite} we describe how a call site is represented in the
model; this encoding is performed similarly across architectures. We use the
same encoding in three different popular architectures - LSTM-based attention
encoder-decoder (\cref{SSec@Model@LSTM}), a Transformer
(\cref{SSec@Model@Transformer}), and a GNN (\cref{SSec@Model@GNNs}). \footnote{We do not share any learnable
parameters nor weights across architectures; every architecture is trained
from scratch.}

\subsection{Call Site Encoder}\label{SSec@Model@Callsite}
We define a vocabulary of learned embeddings $E^{names}$. This vocabulary assigns a vector for every \emph{sub}token of an API name that was observed in the
training corpus. For example, if the training corpus contains a call to
\scode{open\_file}, each of \scode{open} and \scode{file} is assigned a
vector in $E^{names}$.

Additionally, we define a learned embedding for each abstract value,
e.g., ARG, STK, RET and GLOBAL (\cref{Sec@Repr}), and for every actual value (e.g., the number ``\scode{1}'') that occurred in training data.
We denote the
matrix containing these vectors as $E^{values}$. We represent a call site by summing
the embeddings of its API subtokens $w_1...w_{k_s}$, and concatenating with up to $k_{args}$
of argument abstract value embeddings:
\begin{equation*}
encode\_callsite  \left(w_1...w_{k_{s}},value_1,...,value_{k_{args}}\right)= 
\left[\left(\sum_{i}^{k_{s}}E^{names}_{w_i}\right) \; ; \; E^{values}_{value_1} \; ; \; ... \; ; \; E^{values}_{{value_{k_{args}}}} \right] 
\end{equation*}
We pad the remaining of the $k_{args}$ argument slots with an additional \scode{no-arg} symbol. %

\subsection{Encoding Call Site Sequences with LSTMs}\label{SSec@Model@LSTM}
In LSTMs and Transformers, we extract all paths between $Entry$ and $Sink$ (\cref{Sec@Repr}) and use this set of paths to represent the procedure.
We follow the general encoder-decoder paradigm \cite{cho2014learning, sutskever2014sequence} for sequence-to-sequence (seq2seq) models, with the
difference being that the input is not the standard \emph{single} sequence of symbols, but a
\emph{set} of call site sequences. Thus, these models can be described as ``set-of-sequences-to-sequence''.
We learn sequences of encoded call sites; finally, we decode the target procedure name word-by-word while considering a
dynamic weighted average of encoded call site sequence vectors at each step.

In the LSTM-based model, we learn a call site sequence using a bidirectional LSTM. We represent each call site sequence by concatenating the last states of forward
and backward LSTMs:
\begin{equation*}
h_1, ...,h_{l} = LSTM_{encoder}\left(callsite_{1},...,callsite_{l}\right) \nonumber
\end{equation*}
\begin{equation*}
z=\left[h_{l}^{\rightarrow};h_{l}^{\leftarrow}\right] \nonumber
\end{equation*}

Where $l$ is the maximal length of a call site sequence. In our experiments, we used $l=60$. Finally, we represent
the entire procedure as a set of its encoded call site sequences 
$\left\{z_1,z_2,...,z_n\right\}$, which is passed to the LSTM decoder as described in \cref{SSec@Background@attention}.

\subsection{Encoding Call Site Sequences with Transformers}\label{SSec@Model@Transformer}
In the Transformer model, we encode each call site sequence separately using $N$ Transformer-encoder layers, followed by attention pooling to represent a call site sequence as a single vector:
\begin{equation*}
z = \text{Transformer}_{encoder}\left(callsite_{1},...,callsite_{l}\right) 
\end{equation*}
Finally, we represent
the entire procedure as a set of its encoded call site sequences 
$\left\{z_1,z_2,...,z_n\right\}$, which is passed to the Transformer-decoder as described in \cref{SSec@Background@transformers}.

\subsection{Encoding Call Site Sequences with Graph Neural Networks}\label{SSec@Model@GNNs}

In the GNN model, we model the entire processed \ac{CFG} as a graph. We learn
the graph using a \ac{GCN} \cite{kipf2017semi}. 

We begin with a graph that is identical to the \ac{CFG}, where nodes are
\ac{CFG} basic blocks. A basic block in the \ac{CFG} contains multiple call
sites. We thus split each basic block to a chain of its call sites, such that
all incoming edges are connected to the first call site, and the last call
site is connected to all the basic block's outgoing edges.

In a given call site in our analysis, the \emph{concrete and abstract values} (\Cref{Sec@Repr}) depend on the CFG
path that ``arrives'' to the call site.
In other words, the same call site argument can be assigned two different
abstract values in two different runtime paths.  We thus duplicate each call
site that may have different combination of abstract values into multiple
parallel call sites. All these duplicates contain the same API call, but with
a different combination of possible abstract values. Note that while these
values were deduced in the context of simple paths extracted from the
\ac{CFG}, here they are placed inside a graph which contains loops and other
non-simple paths.

%% file: eval.tex
\section{Evaluation}\label{Sec@Eval}
We implemented our approach in a framework called \tool, for \emph{NEural Reverse engineering Of stripped binaries}.

\subsection{Experimental Setup} \label{SSec@Eval@TestsetCreation}

To demonstrate how our representation can be plugged into various
architectures, we implemented our approach in three models: \emph{\toolL}
encodes the control-flow sequences using bidirectional \acp{LSTM} and decodes with
another \ac{LSTM};  \emph{\toolT} encodes these sequences and decodes using a
Transformer \cite{vaswani2017attention}; \emph{\toolG} encodes the \ac{CFG} of
augmented call sites using a \ac{GCN}, and attends to the final node
representations while decoding using an \ac{LSTM}.

The neural architecture of \toolL{} is similar to the architecture of \emph{code2seq} \cite{alon2018code2seq}, with the main difference that \toolL{} is based on sequences of augmented call sites, while \emph{code2seq} uses sequences of AST nodes.

\para{Creating a dataset}  We focus our evaluation on Intel 64-bit
executables running on Linux, but the same process can be applied to other
architectures and operating systems. We collected a dataset of software
packages from the GNU code repository containing a variety of applications
such as networking, administrative tools, and libraries. 

To avoid dealing with mixed naming schemes, we removed all packages containing
a mix of programming languages, e.g., a Python package containing partial C
implementations. 
\ignore{We filtered out wrapper procedures because they are usually
very easy to predict and manually reverse-engineer, thus falsely improve the
scores. }

\para{Avoiding duplicates}  Following \citet{lopes2017dejavu} and
\citet{allamanis2018adverse}, who pointed out the existence of code duplication
in open-source datasets and its adverse effects, we created the train,
validation and test sets from completely separate projects and packages.
Additionally, we put much effort, both manual and automatic, into
filtering duplicates from our dataset. To filter duplicates, we filtered out
the following: \begin{enumerate}
	\item \emph{Different versions of the same package} -- for example, ``wget-1.7'' and ``wget-1.20''.
	\item \emph{C++ code} -- C++ code regularly contains overloaded procedures; further, class methods start with the class name as a prefix. To avoid duplication and name leakage, we filtered out all C++ executables entirely.
	\item \emph{Tests} -- all executables suspected as being tests or examples were filtered out.
	\item \emph{Static linking} -- we took only packages that could compile without static linking. This ensures that dependencies are not compiled into the dependent executable.
\end{enumerate}

\para{Detecting procedure boundaries}
Executables are composed from sections, and, in most cases, all code is
placed in one of these sections, dubbed the code section\footnote{This section is
usually named ``text'' in Linux executables}. Stripped executables do not
contain information regarding the placement of procedures from the source-code
in the code section. Moreover, in optimized code, procedures can be in-lined
or split into chunks and scattered across the code section. Executable static
analysis tools employ a linear sweep or recursive descent to detect procedure
boundaries. 

\citet{shin2015recognizing,BYTEWEIGHT2014} show that modern executable static
analysis tools achieve very high accuracy (\textasciitilde 94\%) in detecting
procedure boundaries, yet some misses exist. While we consider detecting
procedure boundaries as an orthogonal problem to procedure
name prediction, we wanted our evaluation to reflect the real-world scenario.
With this in mind, we employed \citeauthor{IDAPRO}, which uses recursive
descent, as our initial method of detecting procedure boundaries. Then, during
our process of analyzing the procedure towards extracting our augmented
call-site based representation, we perform a secondary check of each procedure
to make sure data-flow paths and treatment of stack frames are coherent. This
process revealed some errors in \citeauthor{IDAPRO}'s boundary detection.
Some were able to be fixed while others were removed from our dataset. One example of
such error, which could be detected and fixed, is a procedure calling
\scode{exit()}. In this case, no stack clean-up or \scode{ret} instruction is
put after the call, and the following instructions belong to another
procedure. We note that without debug information, in-lined procedures are not detected
and thus cannot have their name predicted. 

Comparing the boundaries that are automatically detected by this process with
the debug information generated during compilation -- showed that less than
$1\%$ of the procedures suffered from a boundary detection mistake. These
errors resulted in short pieces of code that were moved from one procedure
into another or causing a new procedure to be created. Following our desire to
simulate a real-world scenario we kept all of these procedures in our
dataset.

\para{Obfuscating API names}
A common anti-\ac{RE} technique is obfuscation of identities of APIs
being used by the executable. Tools that implement this technique
usually perform the following: \begin{enumerate}

\item Modify the information that is stored inside sections of the executable in order
to break the connections between calls targeting imported procedures and the
imported procedure.

\item Add code in the beginning of the executable's execution to mimic the
\ac{OS} loader and re-create these connections at runtime. 

\end{enumerate}

The obstructive actions of the first step ((1)) also affect automatic static
analysis tools such as ours. Specifically, the process of reconstructing call
sites for external calls (described in \cref{Sec@Repr}) will be disrupted.
While the number of arguments can still be computed by analyzing  the calling
procedure, the API names can not be resolved, as the semantics of the call and
argument preparation must remain intact. 

In our evaluation, we wanted to test the performance of our
approach in the presence of such API name obfuscation.  
To simulate API name obfuscation tools we removed the most important information used by the dynamic
loader, the name of (or path to) the files to load, and which of the procedures in
them are called by the executable. All of this information is stored in
the \scode{.dynstr} section, as a list of null-character separated strings.
Replacing the content of this section with nulls (zeros) will hinder its
ability to run, yet we made sure that after these manipulations \tools would still be able to analyze the executable correctly.

\para{Dataset processing and statistics} After applying all filtering, our
dataset contains $67,246$ examples. We extracted procedure names to use as
labels, and then created two datasets by: (i) stripping, and (ii) stripping
and obfuscating API names (the API name obfuscation is described
above.)

We split both datasets into the same training-validation-test sets using a
$(8:1:1)$ ratio resulting in 60403/2034/4809 training/validation/test
examples\footnote{The split to 8:1:1 was performed \emph{at the package
level}, to ensure that the examples from each package go to either training OR
test OR validation. The final ratio of the examples turned out to be different
as the number of examples in each package varies.}. Each dataset contains 
$2.42 \left(\pm 0.04\right)$ target
symbols per example. There are $\textbf{489.63} \left(\pm 2.82\right)$
\emph{assembly code tokens} per procedure, which our analysis reduces to
$5.6 \left(\pm 0.03\right)$ nodes (in GNN);
 $9.05 \left(\pm 0.18\right)$ paths per procedure (in LSTMs and Transformers);
the average path is $9.15 \left(\pm 0.01\right)$ call sites long. 
Our datasets are publicly available at \oursite.

Procedures that do not have any API calls, but still have \emph{internal}
calls, can still benefit from our augmented call site representation. These
constitute 15\% of our benchmark.

\ignore{

While call sites to external procedures, which contain API names, are more
useful, as we show in \cref{Fig@Repr@ACSG_ext} call sites for internal
procedures are included in our represenatio as well. 

}

\para{Metrics} At test and validation time, we adopted the measure used by
previous work \cite{allamanis16convolutional, alon2018code2seq,
fernandes2018structured}, and measured precision, recall and F1 score over the
target \emph{sub}tokens, case-, order-, and duplication-insensitive,
and ignoring non-alphabetical characters. For example, for a true reference of
\scode{open file}: a prediction of \scode{open} is given full precision and
$50\%$ recall; and a prediction of \scode {file open input file} is given
$67\%$ precision and full recall.

\para{Baselines} We compare our models to the state-of-the-art, non-neural, model \debins
\cite{he2018debin}. This is a non-neural baseline based on Conditional Random
Fields (CRFs). We note that \debins was designed for a slightly different task
of predicting names for both local variables and procedure names.
Nevertheless, we focus on the task of predicting procedure names and use only
these to compute their score. 
The authors of \debin{} \cite{he2018debin} offered their guidance and advice in performing a fair comparison, and they provided additional evaluation scripts.

 \dires \cite{lacomis2019dire} is a more recent work, which creates a
representation for binary procedure elements based on \citeauthor{HexRays}, a black-box
commercial decompiler. One part of the representation is based on a textual
output of the decompiler -- a sequence of C code tokens, that is fed to an
\ac{LSTM}. The other part is an \ac{AST} of the decompiled C code,
created by the decompiler as well, that is fed into a GNN.
\dire{} trains the \ac{LSTM} and the \ac{GNN} jointly.
\dire{} addresses a task related 
to ours -- predicting names for local variables.
We therefore adapted \dire{} to our task of predicting procedure names by modifying the authors' original code to predict procedure names only. %

We have trained and tested the \debins and \dires models on our dataset.\footnote{The
dataset of \debin{} is not publicly available. \dires only made their
generated procedure representations available, and thus does not allow to
train a different model on the same dataset. We make our original executables
along with the generated procedure representations dataset public at \oursite.} As in
\tools, we verified that our API name obfuscation method does prevent \debins
and \dire{} from analyzing the executables correctly.

Other straightforward baselines are \emph{Transformer-text} and
\emph{LSTM-text} in which we do not perform \emph{any} program analysis, and
instead just apply standard \ac{NMT} architectures directly on the assembly code:
one is the Transformer which uses the default hyperparameters of \citet{vaswani2017attention}, and the other has two
layers of bidirectional LSTMs with 512 units as the encoder, two LSTM layers with 512 units in the decoder, and
attention \cite{luong15}.

\para{Training} We trained our models using a single Tesla V100 GPU. For all
our models (\toolL{},  \toolT{} and \toolG{}), we used embeddings of size $128$
for target subtokens and API subtokens, and the same size for embedding
argument abstract values. In our LSTM model, to encode call site sequences, we
used bidirectional \acp{LSTM} with $128$ units each; the decoder LSTM had $256$
units. We used dropout \cite{srivastava2014dropout} of $0.5$ on the embeddings
and the LSTMs. For our Transformer model we used $N=6$ encoder layers and the
same number of decoder layers, keys and values of size $d_k=d_v=32$, and a
feed-forward network of size $128$. 
In our \ac{GCN} model we used 4 layers, by
optimizing this value on the validation set; larger values showed only minor
improvement. For all models, we used the Adam \cite{kingma2014adam}
optimization algorithm with an initial learning rate of $10^{-4}$ decayed by a
factor of $0.95$ every epoch. We trained each network end-to-end using the
cross-entropy loss. We tuned hyperparameters on the validation set, and
evaluated the final model on the test set.

\input{tables/results} 

\subsection{Results}\label{SSec@Eval@Debin}

The left side of \cref{TAB@Eval@DebinTAB} shows the results of the comparison
to \debin{}, \dire, \emph{LSTM-text}, and \emph{Transformer-text} on the
stripped dataset. 
Overall, our models outperform all the baselines. This shows
the usefulness of our representation with different learning architectures. 

\toolT{} performs similarly to \toolL{}, while \toolG{} performs better than
both. \toolG{} show $\NeroVSDIRE$ relative improvement
over \dire, $\NeroVSDebin$ over \debin, and over $100\%$ relative gain over \emph{LSTM-text} and \emph{Transformer-text}.

\ignore{
Comparing the models that employ static analysis, which are all \tools models and
\debin{}, to the other models that train on raw disassembly or decompilation outputs  -- 
shows the importance of combining static analysis with neural approaches in this challenging domain of stripped binaries.}

The right side of \cref{TAB@Eval@DebinTAB} shows the same models on the
\emph{stripped and API-obfuscated} dataset. Obfuscation degrades the results
of all models, yet our models still outperform \dires, \debin{}, and the
textual baselines. These results depict the importance of augmenting call
sites using abstract and concrete values in our representation, which can
recover sufficient information even in the absence of API names. We note that
overall, in both datasets, our models perform best on both precision, recall,
and F1.

\para{Comparison to \debins} Conceptually, our model is much more
powerful than the model of \debins because it is able to decode
out-of-vocabulary procedure names (``neologisms'') from subtokens, while the
CRF of \debins uses a closed vocabulary that can only predict
already-seen procedure names.  At the binary code side, since our model is
neural, at test time it can utilize unseen call site sequences while their CRF
can only use observed relationships between elements. For a detailed
discussion about the advantages of neural models over CRFs, see Section 5 of
\cite{alon2019code2vec}. Furthermore, their representation performs a shallow
translation from binary instruction to connections between symbols, while our
representation is based on a deeper data-flow-based  analysis to find values
of registers arguments of imported procedures. 

\para{Comparison to \dires} The \dire{} model and our \toolG{} model use similar neural
architectures; yet, our models perform much better. While decompilation
evolves static analysis, its goal is to generate \emph{human readable} output.
On the other hand, our representation was tailor made for the prediction task,
by focusing on the call sites and augmenting them to encode more information.

\para{Comparison to LSTM-text and Transformer-text} The comparison to the
\ac{NMT} baselines shows that learning directly from the assembly code
performs significantly worse than leveraging semantic knowledge and static
analysis of binaries. We hypothesize that the reasons are the high variability
in the assembly data, which results in a low signal-to-noise ratio. This
comparison necessitates the need of an informative static analysis to
represent and learn from executables.

\input{eval_examples}

\para{Examples} \Cref{TAB@Eval@Examples} shows a few examples for predictions
made by the different models. Additional examples can be found in \cref{appendix}.

\subsection{Ablation study} \label{SSec@Eval@Ablation}

To evaluate the contribution of different components of our representation, we compare  the following configurations:

\textbf{\emph{\toolL{} no values}} - uses only the CFG analysis with the called API \emph{names}, without abstract nor concrete values, and uses the LSTM-based model. 

\textbf{\emph{\toolT{} no values}} - uses only the CFG analysis with the called API \emph{names}, without abstract nor concrete values, and uses the Transformer-based model. 

\textbf{\emph{\toolG{} no-values}} - uses only the CFG analysis without abstract nor concrete values, and uses the GNN-based model.

\textbf{\emph{\toolL{} no-library-debug}} - does not use debug information for external dependencies when reconstructing call sites\footnote{To calculate the number of arguments in external calls only caller and callee code is used}, and uses the LSTM-based model. 

\textbf{\emph{\toolT{} no-library-debug}} - does not use debug information for external dependencies when reconstructing call sites, and uses the Transformer-based model. 

\textbf{\emph{\toolG{} no-library-debug}} - does not use debug information for external dependencies when reconstructing call sites, and uses the GNN-based model.

\textbf{\emph{\toolsTL}} -  uses a Transformer to encode the sets of
control-flow sequences and an LSTM to decode the prediction.

\textbf{\emph{BiLSTM call sites}} -  uses the same enriched call sites
representation as our model including abstract values, with the main difference being
that the order of the call sites is \emph{their order in the assembly code}:
there is no analysis of the CFG.

\textbf{\emph{BiLSTM calls}} - does not use CFG analysis or abstract values.
Instead, it uses two layers of bidirectional LSTMs with attention to encode
\scode{call} instructions with only the name of the called procedure, in the
order they appear in the executable. 

\para{Results} \Cref{TAB@Eval@AblationTAB} shows the performance of the
different configurations.  \toolL{} achieves ~$50\%$ higher relative score
than \emph{\toolL{} no-values}; \toolG{} achieves $20\%$ higher relative score than \emph{\toolG{} no-values}. This shows the contribution of the call site
augmentation by capturing values and abstract values and its importance to
prediction. 

The \emph{no-library-debug} variations of each model achieve comparable
results to the originals (which using the debug information from the
libraries). These results showcase the precision of our analysis and its
applicability to possible real-world scenarios in which library code is
present without debug information.

\toolsTL{}~achieved slightly lower results to \toolT{} and \toolL{}. This
shows that the representation and the information that is captured in it --
affect the results much more than the specific neural architecture.

As we discuss in \cref{Sec@Repr}, our data-flow-based analysis helps filtering
and reordering calls in their approximate chronological runtime order, rather
than the arbitrary order of calls as they appear in the assembly code.
\emph{BiLSTM call sites} performs slightly better than \emph{BiLSTM calls} due
to the use of abstract values instead of plain \scode{call} instructions.
\toolL{} improves over \emph{BiLSTM call sites} by 16\%, showing the
importance of learning call sites \emph{in their right chronological order}
and the importance of our data-flow-based observation of executables.

\input{tables/ablation}

\subsection{Qualitative evaluation} Taking a closer look at partially-correct
predictions reveals some common patterns. We divide these partially-correct predictions into
major groups; \cref{SSec@Eval@Errors@table} shows a few examples from interesting and common groups.

The first group, ``Programmers VS English Language'', depicts prediction
errors that are caused by programmers' naming habits. The first two
lines of \cref{SSec@Eval@Errors@table} show an example of a relatively common practice for programmers -- the
use of shorthands. In the first example, the ground truth is
\scode{i18n\_initialize}\footnote{Internationalization (i18n) is the process
of developing products in such a way that they can be localized for languages
and cultures easily.}; yet the model predicts the shorthand \scode{init}. In
the second example, the ground truth subtoken is the shorthand \scode{cfg} but the model
predicts \scode{config}. We note that \scode{cfg}, \scode{config} and \scode{init} appear 
$81$, $496$ and $6808$ times, respectively, in our training set. However, \scode{initialize} does not appear at all, thus justifying the model's
prediction of \scode{init}.

In the second group, ``Data Structure Name Missing'', the model, not privy to 
application-specific labels that exist only in the test set, resorts to
predicting generic data structures names (\ie a list). That is, instead of predicting the application-specific structure, the model predicts the most similar generic data structure in terms of common memory and API use-patterns. In the first example, \scode{speed} records are
predicted as \scode{list items}. In the next example, the parsing of a (windows)
directories, performed by the \scode{wget} package, is predicted as parsing a
\scode{tree}. In the last example, the subtoken \scode{gzip} is out-of-vocabulary, causing \scode{abort\_gzip} to be predicted instead as \scode{fatal}. 

The last group, ``Verb Replacement'', shows how much information is captured
in \emph{verbs} of procedure name. In the first example, \scode{share} is replaced
with \scode{add}, losing some information about the work performed in the
procedure. In the last example, \scode{display} is replaced with \scode{show}, which are
almost synonymous.

\ignore{

The last group, ``Literal Name'', shows what happens when the procedure
performs some high-level function, while the model's prediction only provides
a literal description of what happens in the procedure. The first line shows
how the high-level concept of activating a usb device (in the case of libpcap to
allow usb sniffing) is missed by the model and instead predicted as a literal
operation of logging (the operation), adding (the record) and opening (the
socket to the usb interface).

}

\input{tables/quality}

\subsection{Limitations} 
In this section, we focus on the limitations in our approach, which could serve as a basis for future research.

\para{Heavy obfuscations and packers}  In our evaluation, we explored the
effects of API obfuscation and show that they cause a noticeable decrease in
precision for all name prediction models reviewed (\cref{SSec@Eval@Debin}). 
Alternatively, there are more types of obfuscators, packers, and other types
of self-modifying code. 
Addressing these cases better involves detecting them
at the static analysis phase and employing other methods (\eg
\citet{PolyUnpackA}) before attempting to predict procedure names.

\para{Representation depending on call sites}  As our representation for procedures
is based on call site reconstruction and augmentation, it is dependent on the
procedure having at least one call site and the ability to detect them
statically.

Moreover, as we mentioned above, $15\%$ of the procedures in our dataset
contain only internal calls. These are small procedures with an average of 4
(internal or indirect) calls and 10 basic blocks. These procedures are
represented only by call sites of internal and indirect calls. These call
sites are composed of a special ``UNKNOWN'' symbol with the abstract or
concrete values. Predictions for these procedures receive slightly higher
results - $46.26$, $49.54$, $47.84$ (Precision, Recall, F1). This is consistent
with our general observation that smaller procedures are easier to predict. 

\para{Predicting names for C++ or python extension modules}  There is no
inherent limitation in Nero preventing it from predicting class or class
method names. In fact, training on C++ code only and incorporating more
information available in C++ compiled binaries (as shown in \cite{KatzClass})
in our representation makes for great material for future work. Python
extension modules contain even more information as the python types are
created and manipulated by calling to the python C\\C++ API (\eg
\scode{\_PyString\_FromString}). 

We decided to focus on binaries created from C code because they contain the
\emph{smallest} amount of information and because this was also done in other works that we compared against
 (\citet{he2018debin,lacomis2019dire}).

%% file: tables/results.tex
\begin{table*}[t]
\vspace{-2mm}
\centering
\begin{tabular}{@{}llllllll@{}}
\toprule
& \multicolumn{3}{c}{Stripped} & & \multicolumn{3}{c}{\emph{Stripped \& Obfuscated API calls}} \\
\cmidrule{2-4} \cmidrule{6-8} 
Model               			& Precision & Recall 	& F1 				& & Precision 	& Recall 	& F1   \\
\midrule
LSTM-text 						& 22.32    	& 21.16  	& \ScoreLSTM 		& & 15.46 		& 14.00 	& 14.70 \\
Transformer-text    			& 25.45    	& 15.97  	& 19.64 	 		& & 18.41 		& 12.24 	& 14.70 \\
\debin{} \cite{he2018debin}  	& 34.86    	& 32.54  	& \ScoreDebin 		& & 32.10 		& 28.76 	& 30.09 \\
\dire{} \cite{lacomis2019dire}  & 38.02    	& 33.33  	& 35.52 	  		& & 23.14 		& 25.88 	& 24.43  \\
\midrule
\toolL   						& 39.94     & 38.89     & 39.40      		& & 39.12 		& 31.40     & 34.83 \\
\toolT   						& 41.54     & 38.64     & 40.04 			& & 36.50 		& 32.25 	& 34.24      \\
\toolG   						& \BF{48.61}& \BF{42.82}& \BF{\ScoreGNero} 	& & \BF{40.53} 	& \BF{37.26}& \BF{38.83} \\
\bottomrule
\end{tabular}
\caption{Our models outperform previous work, \dires and \debins, by a relative improvement of $\NeroVSDIRE$ and $\NeroVSDebin$ resp.; learning from the flat assembly code (LSTM-text, Transformer-text) yields much lower results. \emph{Obfuscating API calls} hurts all models, but thanks to the use of \emph{abstract and concrete values}, our model still performs better than the baselines.}
\label{TAB@Eval@DebinTAB}
\vspace{-3mm}
\end{table*}

%% file: eval_examples.tex
\begin{table*}[t]
\centering
\normalsize
\begin{tabular}{@{}lllll@{}}
\toprule
Model					& \multicolumn{4}{c}{Prediction}  \\ 
\midrule[\heavyrulewidth]
Ground truth				& locate unset 			& free words	 		 & get user groups 			& install signal handlers\\
\midrule[\heavyrulewidth]
\debin 						& var is \textbf{unset}	& search 	 			 & display 					&  \textbf{signal} setup \\ 
\dire  						& env concat			& restore		 		 & prcess file 				&  overflow              \\ 
\emph{LSTM-text}			& url get arg 			& func \textbf{free} 	 & <unk>					&  <unk>				 \\
\emph{Transformer-text} 	& <unk>					& <unk>				 	 & close stdin				&  <unk>				 \\
\midrule[\heavyrulewidth]
\toolL 						& var is \textbf{unset} & quotearg \textbf{free} & \textbf{get user groups} & enable mouse 		 	 \\
\toolT						& var is \textbf{unset} & quotearg \textbf{free} & open op 					& <empty> 				 \\
\toolG 						& var is \textbf{unset} & \textbf{free} table	 & \textbf{get user groups} & \textbf{signal} enter \textbf{handlers} \\
\bottomrule
\end{tabular}
\caption{Examples from our test set and predictions made by the different models. Even when a prediction is not an ``exact match'' to the ground truth, it usually captures more subtokens of the ground truth than the baselines. More examples can be found in \cref{appendix}.}
\label{TAB@Eval@Examples}
\end{table*}

%% file: tables/ablation.tex
\begin{table}[H]
\centering
\normalsize
\begin{tabular}{@{}llll@{}}
\toprule
Model                 & Prec    & Rec     & F1    \\
\midrule
BiLSTM calls          & 23.45   & 24.56   & 24.04 \\
BiLSTM call sites 	  & 36.05   & 31.77   & 33.77 \\
\toolL{} no-values    & 27.22   & 23.91   & 25.46 \\
\toolT{} no-values    & 29.84   & 24.08   & 26.65 \\
\toolG{} no-values    & 45.20   & 32.65   & 37.91 \\
\toolL{} no-library-debug    & 39.51   & 40.33   & 39.92 \\
\toolT{} no-library-debug    & 43.60   & 37.65   & 40.44 \\
\toolG{} no-library-debug    & 47.73   & 42.82   & 45.14 \\
\toolsTL{}            & 39.05   & 36.47   & 37.72 \\
\midrule
\toolL{}       		  & 39.94   & 38.89   & 39.40 \\
\toolT{}              & 41.54   & 38.64   & 40.04 \\
\toolG{}              & 48.61   & 42.82   & \ScoreGNero \\
\bottomrule
\end{tabular}
\caption{Variations on our models that ablate different components.}
\label{TAB@Eval@AblationTAB}
\vspace{-10pt}
\end{table}

%% file: tables/quality.tex
\begin{table*}[h]
\begin{tabular}{cccc}
\toprule
\multicolumn{1}{c}{Error Type}                                                                                & \multicolumn{1}{c}{Package}  & \multicolumn{1}{c}{Ground Truth}      & \multicolumn{1}{c}{Predicted Name}      \\ \midrule
\multicolumn{1}{c}{\multirow{3}{*}{\begin{tabular}[c]{@{}c@{}}Programmers VS\\ English Language\end{tabular}}} & \multicolumn{1}{c}{wget}     & \multicolumn{1}{c}{i18n\_initialize} & \multicolumn{1}{c}{i18n\_init}          \\  
\multicolumn{1}{c}{}                                                                                           & \multicolumn{1}{c}{direvent} & \multicolumn{1}{c}{split\_cfg\_path} & \multicolumn{1}{c}{split\_config\_path} \\ 
\multicolumn{1}{c}{}                                                                                           & gzip                          & add\_env\_opt                         & add\_option                             \\ \midrule
\multirow{4}{*}{\begin{tabular}[c]{@{}c@{}}Data Structure \\ Name Missing\end{tabular}}                         & gtypist                       & get\_best\_speed                      & get\_list\_item                         \\ 
                                                                                                                & wget                          & ftp\_parse\_winnt\_ls                 & parse\_tree                             \\
                                                                                                                & direvent&filename\_pattern\_free&free\_buffer\\  
                                                                                                                & gzip                          & abort\_gzip\_signal                   & fatal\_signal\_handler                  \\ \midrule
\multicolumn{1}{c}{\multirow{4}{*}{\begin{tabular}[c]{@{}c@{}}Verb Replacement\end{tabular}}}                   & findutils                     & share\_file\_fopen                    & add\_file                               \\
\multicolumn{1}{c}{}                                                                                            & units                         & read\_units                           & parse                                  \\ 
\multicolumn{1}{c}{}                                                                                            & wget                          & retrieve\_from\_file                  & get\_from\_file                         \\  
\multicolumn{1}{c}{}                                                                                            & mcsim                         & display\_help                         & show\_help                              \\ 
\bottomrule
\end{tabular}
\caption{Examination of common interesting model mistakes.}
\label{SSec@Eval@Errors@table}
\vspace{-10pt}
\end{table*}

%% file: related.tex
\section{Related Work}\label{Sec@Related}

\ignore{
\para{Name prediction in decompiled code}
\begin{itemize}
	\item https://dl.acm.org/citation.cfm?id=3121274
\end{itemize}

\para{Name prediction in Byte-code}
\begin{itemize}
	\item http://www.duboue.net/pablo/papers/ASAI2018choice.pdf
\end{itemize}
Another work used Conditional Random Fields (CRFs) to name prediction in bytecode \cite{android2016}.
}

\para{Machine learning for source code}
Several works have investigated machine learning approaches for predicting names in high-level languages. Most works focused on variable names \cite{alon2018general, bavishi2018context2name}, method names \cite{allamanis16convolutional, alon2019code2vec, allamanis2015} or general properties of code \cite{raychev2016noisy, raychev14}. Another interesting application is measuring the likelihood of existing names to detect naming bugs \cite{pradel2018deepbugs, rice2017detecting}. Most work in this field used either syntax only \cite{bielik16phog, decisionTrees2016, maddison2014structured}, semantic analysis \cite{allamanis2017learning} or both \cite{raychev2015jsnice, iyer2018mapping}.
Leveraging syntax \emph{only} may be useful in languages such as Java and JavaScript that have a rich syntax, which is not available in our difficult scenario of \ac{RE} of binaries. 
In contrast with syntactic-only work such as \citet{alon2018code2seq, alon2019code2vec}, working with binaries requires a deeper semantic analysis in the spirit of \citet{allamanis2017learning}, which recovers sufficient information for training the model using semantic analysis. 

\citet{allamanis2017learning}, \citet{brockschmidt2018generative} and \citet{fernandes2018structured} further leveraged semantic analysis with \acp{GNN}, where edges in the graph were relations found using syntactic and semantic analysis. 
Another work \cite{defreez2018} learned embeddings for C functions based on the \ac{CFG}. We also use the \ac{CFG}, but in the more difficult domain of stripped \emph{compiled} binaries rather than C code.

\para{Static analysis models for \ac{RE}} Debin \cite{he2018debin} used static analysis with CRFs to predict various properties in binaries. As we show in \cref{Sec@Eval}, our model gains $20\%$ higher accuracy due to Debin's sparse model and our deeper data-flow analysis. 

In a concurrent work with ours, DIRE \cite{lacomis2019dire}  addresses name
prediction for local variables in binary procedures.  Their approach relies on
a commercial decompiler and its evaluation was performed only on non-stripped
executables. Our approach works directly on the optimized binaries.
Moreover,
they focused on predicting variable names, which is more of a local prediction;
in contrast, we focus on predicting procedure names, which requires a more global view
of the given binary.  Their approach combines an LSTM applied on the flat
assembly code with a graph neural network applied on the AST of the decompiled
C code, without a deep static analysis as in ours and without leveraging the CFG.
Their approach is similar to the \emph{LSTM-text} baseline combined with
the \toolL{} no-values ablation. As we show in \cref{Sec@Eval}, our GNN-based
model achieves $\NeroVSDIRE$ higher scores thanks to our deeper data-flow analysis.

\para{Static analysis frameworks for \ac{RE}} \citet{KatzClass} showed an approach to infer subclass-superclass relations in stripped binaries. \citet{TIE} used static and \emph{dynamic} analysis to recover high-level types. In contrast, our approach is purely \emph{static}.
\cite{Surfer} presented CodeSurfer, a binary executable framework built for analyzing x86 executables, focusing on detecting and manipulating variables in the assembly code. 
\citet{shin2015recognizing,BYTEWEIGHT2014} used RNNs to identify procedure boundaries inside a stripped binary. 

\para{Similarity in binary procedures} \citet{GITZ} and \citet{Pewny} addressed the problem of finding similar procedures to a given procedure or executable, which is useful to detect vulnerabilities. \citet{xu2017neural} presents Gemini, a \ac{DNN} based model for establishing binary similarity. Gemini works by annotating \ac{CFG} with manually selected features, and using them to create embeddings for basic blocks and create a graph representation. This representation is then fed into a Siamese architecture to generate a similarity label (similar or not similar). \citet{Asm2Vec} propused Asm2Vec. Asm2Vec works by encoding assembly code and the \ac{CFG} into a feature vector and using a PV-DM based model to compute similarity.

\ignore{
\AWK{
\para{Visualizing binary code}
}
}

%% file: conclusion.tex
\section{Conclusion} 

We present a novel approach for predicting procedure names in stripped
binaries. The core idea is to leverage static analysis of binaries to encode
rich representations of API call sites; use the \ac{CFG} to
approximate the chronological runtime order of the call sites, and encode the
CFG using either a set of sequences or a graph, using three different
neural architectures (LSTM-based, Transformer-based, and graph-based).

We evaluated our framework on real-world stripped procedures. Our model
achieves a $\NeroVSDebin$ relative gain over existing non-neural approaches,
and more than a $\NeroVSText$ relative gain over the na\"ive textual baselines
(``LSTM-text'', ``Transformer-text'' and \dire). Our ablation study shows the
importance of analyzing argument values and learning from the CFG. To the best
of our knowledge, this is the first work to leverage deep learning for reverse
engineering procedure names in binary code.

We believe that the principles presented in this paper can serve as a basis
for a wide range of tasks that involve learning models and RE, such as malware
and ransomware detection, executable search, and neural decompilation. To this
end, we make our dataset, code, and trained models publicly
available at \oursite.

%% file: ack.tex
\section*{Acknowledgements} 
We would like to thank Jingxuan He and Martin Vechev for their help in running Debin, and Emery Berger for his useful advice.
We would also like to thank the anonymous reviewers for their useful suggestions.

The research leading to these results has received funding from the Israel
Ministry of Science and Technology, grant no. 3-9779.

%% file: appendix.tex
\section{Additional Examples}
\label{appendix}
\Cref{TAB@Supp@Examples} contains more examples from our test set, along with the predictions made by our model and each of the baselines. %
\renewcommand{\arraystretch}{1.5}

\begin{table}[h!]
\footnotesize
\begin{tabular}{@{}l|lllll@{}}
\toprule
Ground Truth		& \citet{he2018debin}	 & \emph{LSTM-text}	& \emph{Transformer-text} &  BiLSTM call-sites & \toolL{} \\
\midrule

mktime from utc	&nettle pss ...	&get boundary	&<unk>	&str file	&\textbf{mktime} \\
read buffer	&concat	&fopen safer	&mh print fmtspec	&net \textbf{read}	&filter \textbf{read} \\
get widech	&\textbf{get} byte	&user	&mh decode rcpt flag	&<unk>	&do tolower \\
ftp parse winnt ls	&uuconf iv ...	&mktime 	&print status	&send to file	&\textbf{parse} tree \\
write init pos	&allocate pic buf	&open int	&<unk>	&print type	&cfg \textbf{init} \\
wait for proc	&\textbf{wait} subprocess	&start open	&mh print fmtspec	&<unk>	&strip \\
read string	&cmp	&error	&check command	&process 	&io \textbf{read} \\
find env	&\textbf{find} \textbf{env} pos	&proper name utf	&close stream	&read token	&\textbf{find} \textbf{env} \\
write calc jacob	&usage msg	&update pattern	&print one paragraph	&<unk>	&\textbf{write} \\
write calc outputs	&fsquery show	&debug section	&cwd advance fd	&<unk>	&\textbf{write} \\
get script line	&\textbf{get} \textbf{line}	&make dir hier	&<unk>	&read ps \textbf{line}	&jconfig \textbf{get} \\
getuser readline	&stdin read \textbf{readline}	&rushdb print	&mh decode rcpt flag	&write line	&\textbf{readline} read \\
set max db age	&do link	&\textbf{set} owner	&make dir hier	&sparse copy	&\textbf{set} \\
write calc deriv	&orthodox hdy	&ds symbol	&close stream	&fprint entry	&\textbf{write} type \\
read file	&bt open	&<unk>	&... disable coredump	&<unk>	&vfs \textbf{read} \textbf{file} \\
parse options	&\textbf{parse} \textbf{options}	&finish	&mh print fmtspec	&get \textbf{options}	&\textbf{parse} args \\
url free	&hash rehash	&hostname destroy	&setupvariables	&hol \textbf{free}	&\textbf{free} dfa content \\
check new watcher	&read index	&\textbf{check} opt 	&<unk>	&open source	&\textbf{check} file \\
open input file	&get options	&query in 	&ck rename	&set 	&delete \textbf{input} \\
write calc jacob	&put in fp table	&save game var	&hostname destroy	&<unk>	&\textbf{write} \\
filename pattern free	&add char segment	&\textbf{free} dfa content	&hostname destroy	&glob cleanup	&\textbf{free} exclude segment \\
read line	&tartime	&init all	&close stdout	&parse args	&\textbf{read} \\
ftp parse unix ls	&serv select fn	&canonicalize 	&<unk>	&<unk>	&\textbf{parse} syntax option \\
free netrc	&gea compile	&hostname destroy	&hostname destroy	&\textbf{free} ent	&hol \textbf{free} \\
string to bool	&\textbf{string} \textbf{to} \textbf{bool}	&setnonblock	&mh decode rcpt flag	&\textbf{string} \textbf{to} \textbf{bool}	& \textbf{string} \textbf{to} \textbf{bool} \\
\bottomrule
\end{tabular}
\caption{Examples from our test set and predictions made by the different models.}
\label{TAB@Supp@Examples}
\end{table}